\pdfoutput=1

\documentclass[11pt]{article}

\usepackage[]{acl}

\usepackage{times}
\usepackage{latexsym}

\usepackage[T1]{fontenc}

\usepackage[utf8]{inputenc}

\usepackage{microtype}

\usepackage{inconsolata}

\usepackage{graphicx}

\usepackage{enumitem}
\usepackage{todonotes}
\usepackage{algorithm}
\usepackage{algorithmic}
\usepackage{amsfonts}
\usepackage{amsmath}
\usepackage{amssymb}
\usepackage{booktabs}
\usepackage{multirow}
\usepackage{arydshln}
\usepackage{xcolor}
\usepackage{textcomp}
\definecolor{teal2}{HTML}{70C1B3}
\definecolor{blue2}{HTML}{6879FC}
\definecolor{green2}{HTML}{009900}
\definecolor{purple2}{HTML}{6E36CF}

\title{GRAFT: A Graph-based Flow-aware Agentic Framework for Document-level Machine Translation}

\author{
Himanshu Dutta$^{1}$,  
Sunny Manchanda$^{2}$,\\ 
Prakhar Bapat$^{2}$
Meva Ram Gurjar$^{2}$,
Pushpak Bhattacharyya$^{1}$\\
$^{1}$Indian Institute of Technology Bombay, India \\
$^{2}$DYSL-AI, DRDO, India \\
\texttt{\{himanshud, pb\}@cse.iitb.ac.in} \\
\texttt{manchanda.sunny@gmail.com}
}

\begin{document}
\maketitle
\begin{abstract}  

Document-level Machine Translation (DocMT) approaches often struggle with effectively capturing discourse-level phenomena. Existing approaches rely on heuristic rules to segment documents into discourse units, which rarely align with the true discourse structure required for accurate translation. Otherwise, they fail to maintain consistency throughout the document during translation. To address these challenges, we propose \textbf{Gr}aph‐Augmented \textbf{A}gentic \textbf{F}ramework for Document‐Level \textbf{T}ranslation (\textbf{GRAFT}), a novel graph-based DocMT system that leverages Large Language Model (LLM) agents for document translation. Our approach integrates segmentation, directed acyclic graph (DAG) based dependency modelling, and discourse-aware translation into a cohesive framework. Experiments conducted across eight translation directions and six diverse domains demonstrate that GRAFT achieves significant performance gains over state-of-the-art DocMT systems. Specifically, GRAFT delivers an average improvement of 2.8 d-BLEU on the TED test sets from IWSLT2017 over strong baselines and 2.3 d-BLEU for domain-specific translation from English to Chinese. Moreover, our analyses highlight the consistent ability of GRAFT to address discourse-level phenomena, yielding coherent and contextually accurate translations.\footnote{We release our anonymised codebase and data at \texttt{\url{https://anonymous.4open.science/r/graft-docmt/}}.}   
\end{abstract}

\section{Introduction}
\label{sec:introduction}

\begin{figure*}[ht]
    \centering
    \includegraphics[width=0.72\textwidth]{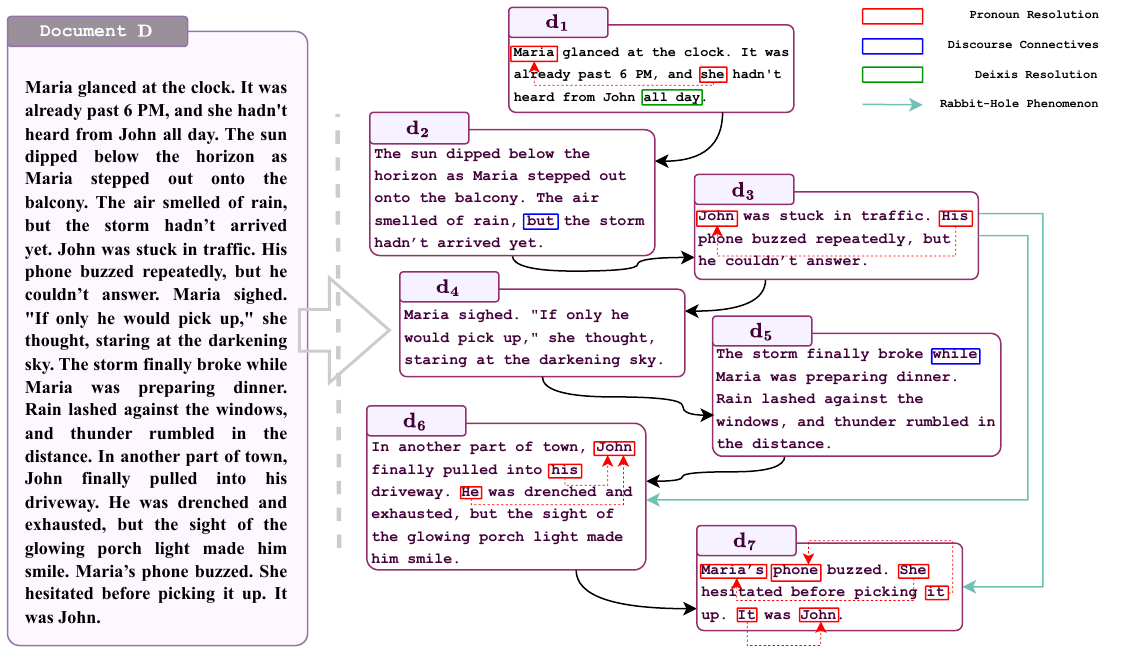}
    \caption{An example illustrating a document \(\mathbf{D}\) segmented into discourse units (\(\mathbf{d_i}\)), represented as nodes in a directed acyclic graph. The structure highlights discourse-level phenomena, including \textbf{sequential connections between discourses}, \textcolor{red}{\textbf{pronoun resolution}}, \textcolor{blue}{\textbf{discourse connectives}}, \textcolor{green2}{\textbf{deixis resolution}}, and \textcolor{teal2}{\textbf{the rabbit-hole phenomenon}}. This representation demonstrates how a document's internal relationships and cohesive elements can be systematically analyzed. These phenomena are invisible to the existing approaches, with varying degrees of effects.}
    \label{fig:example-document}
\end{figure*}

Document‐level Machine Translation (DocMT) seeks to generate translations that not only preserve the meaning of individual sentences but also maintain coherence, cohesion, and consistency across an entire document \cite{bawden-etal-2018-evaluating}. While sentence‐level MT has seen remarkable progress, translating longer texts remains challenging due to discourse phenomena that span multiple sentences or paragraphs \cite{maruf2022survey}.

Discourse phenomena in MT can be classified into \emph{intra‐discoursal} and \emph{inter‐discoursal} phenomena. Intra‐discoursal phenomena, such as pronoun resolution, tense and aspect consistency, ellipsis recovery, and idiomatic expression translation-arise within a coherent segment of text. Inter‐discoursal phenomena, such as entity co‐reference, lexical cohesion (repetition of terminology), discourse connectives, and thematic progression-link separate segments to produce a fluent and coherent document \cite{hu-wan-2023-exploring}.\footnote{We discuss various discourse-level phenomena prominent for the translation task in Appendix~\ref{app:discourse_phenomena}.}

Prior work on document‐level MT may be grouped into several strands. Early methods extended sentence models with context from adjacent sentences via multi‐encoder or concatenation strategies \cite{jean2017does,wang-etal-2017-exploiting-cross}. Hierarchical attention networks condition translation on both word‐ and sentence‐level encodings to capture structured context \cite{miculicich-etal-2018-document}. Cache‐based approaches store recently translated words or topical words in dynamic and topic caches to model coherence \cite{kuang-xiong-2017-modelling, tong-etal-2020-document}. Continuous cache methods leverage a light‐weight history memory to adapt translations on the fly \cite{tu-etal-2018-learning}. More recently, unified context models explicitly encode both local sentence context and global document context in Transformer architectures \cite{ohtani-etal-2019-context}. Large‐scale surveys have summarized these approaches and highlighted persistent gaps in modelling and evaluation \cite{maruf-saleh-haffari-2021-survey}.

\citet{wang2025delta} focuses on treating document-level MT as a sequential translation task. They translate each sentence individually in a sequence, while maintaining a persistent memory, which is updated after the translation of each sentence. Similarly, \citet{hu-wan-2023-exploring} explore the inherent discourse structure present in documents by utilizing the paragraph as the discourse boundary. Both these approaches use heuristic rules to decide discourse boundaries, which are difficult to align with the idea of discourse segmentation for the task of translation.

Ideally, a discourse segmentation strategy should ensure that each discourse unit is self-contained for handling intra-discourse phenomena, such as pronoun resolution and verb tense consistency, utilizing its own context. Simultaneously, it should facilitate the handling of inter-discourse phenomena, like entity co-reference and lexical cohesion, by leveraging context from related discourse units. Further, most memory or cache-based document‐level MT systems discussed above treat context as a flat or heuristic aggregation of preceding sentences, failing to model fine‐grained dependencies between discourse units \citep{bawden-etal-2018-evaluating, voita-etal-2018-context, voita-etal-2019-good}. Figure~\ref{fig:example-document} shows an example document segmented into discourses, annotated with various discourse phenomena. 


These observations reveal three key gaps: (1) Lack of principled discourse segmentation aligned to translation, (2) Absence of structured dependency modelling between discourse units, and (3) Insufficient mechanisms to propagate rich, structured context during translation. To bridge these gaps, we propose \textbf{Graph‐Augmented Translation Framework (GRAFT)}, an agent‐based system for DocMT task that transforms a \emph{source document} $D$ into a directed-acyclic-graph (DAG) structure of the discourses. We introduce four specialized agents: \emph{Discourse Agent}: segments the document into self‐contained discourse units via LLM-based algorithm, each discourse is a node in the DAG structure; \emph{Edge Agent}: establishes directed edges between any pair of discourses whose translations require information‐sharing; \emph{Memory Agent}: extracts structured ``local memory'' (for translation) from each translated discourse; \emph{Translation Agent}: combines each discourse with its incident memories (prioritizing the earliest dependencies) and invokes an LLM to produce contextually informed translations.

\noindent Our contributions are:  
\begin{itemize}[leftmargin=*]
  \item \textbf{Discourse‐Graph Dependency Modelling (Subsection~\ref{subsec:source_document_to_dag}):} We introduce a novel approach to represent the source document for translation as a Directed Acyclic Graph (DAG) to model dependencies between discourse segments effectively. This structural transformation leads to an average performance gain of $\textcolor{blue}{\mathbf{2.0}}$ d-BLEU scores across eight translation directions when compared to translations performed without utilizing the DAG structure (Table~\ref{tab:discourse_agent_ablation} and Table~\ref{tab:edge_agent_ablation}).

  \item \textbf{GRAFT (Section~\ref{sec:methodology}):} We propose a novel LLM agent‐based document-level machine translation system, which segments, connects, and translates discourse units by utilizing an intermediate DAG representation. GRAFT achieves an average performance gain of $\textcolor{blue}{\mathbf{6.4}}$ d-BLEU scores compared to a commercial translation system (\texttt{Google Translate}\footnote{\url{https://translate.google.com/}}), and $\textcolor{blue}{\mathbf{3.4}}$ d-BLEU scores compared to similar LLM-based approaches across eight translation directions (Table~\ref{tab:main_results}). GRAFT achieves comparable performance with large closed-source models (\texttt{OpenAI GPT} models\footnote{\url{https://platform.openai.com/docs/models/gpt-4o-mini}}) while showing qualitative improvements (Table~\ref{tab:main_results} and Table~\ref{tab:domain_results_automatic}).
  
  \item \textbf{Evaluation and Analysis (Section~\ref{sec:experiments_results} and Section~\ref{sec:analysis}):} We evaluate GRAFT on diverse language pairs (\textit{English (en) $\leftrightarrow$ German (de), French (fr), Japanese (ja), Chinese (zh)}) and domains (\textit{Fiction, News, Social, TED, and Novel}). Our analysis highlights the importance of context propagation through the document for translation coherence, with our memory strategies yielding the best results, and reveals that translation consistency based on DAG-based representation aligns better with translation quality.
\end{itemize}

\begin{figure*}[t]
  \centering
  \includegraphics[width=0.74\textwidth]{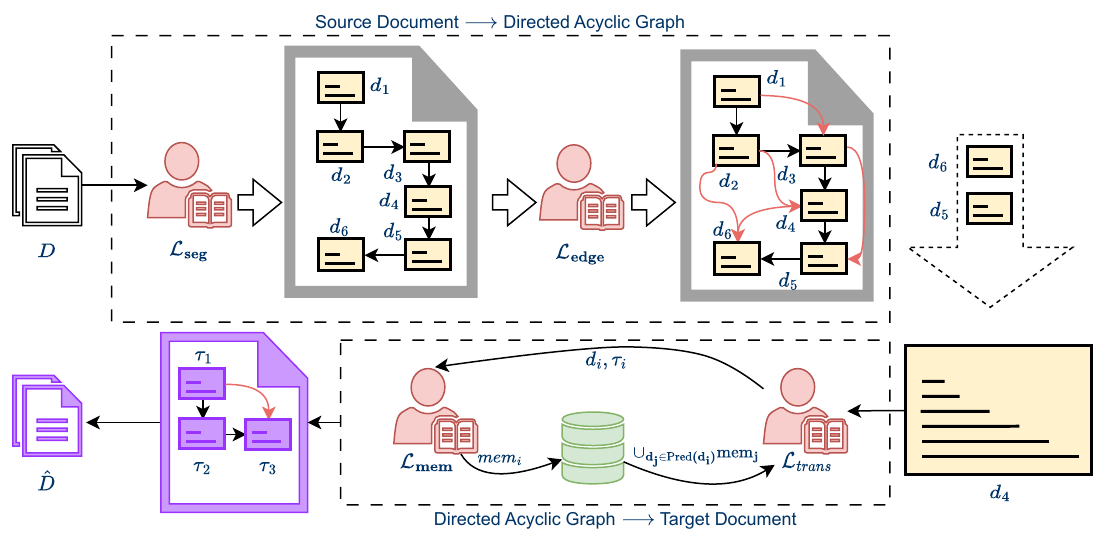}
  \caption{\textbf{GRAFT} pipeline: illustrating the document-level translation pipeline involving four agents: the Discourse Agent, Edge Agent, Translation Agent, and Memory Agent. The process begins with a document $D$ as input, which is segmented into discourse segments/nodes ($d_i$) by the Discourse Agent ($\mathcal{L}_{\mathrm{seg}}$). These segments are then structured into a directed acyclic graph (DAG) with nodes and directed edges, representing dependency relationships, by the Edge Agent ($\mathcal{L}_{\mathrm{edge}}$). The Translation Agent and Memory Agent iteratively process the nodes, where the Translation Agent ($\mathcal{L}_{\mathrm{trans}}$) generates translations for discourse segments using context provided by the Memory Agent ($\mathcal{L}_{\mathrm{mem}}$), which maintains and updates local memory ($\mathrm{mem}_{i}$) based on source and translated discourses. This iterative loop ensures context-aware and cohesive translations, enabling effective handling of intra-discourse phenomena.}
  \label{fig:pipeline_diagram}
\end{figure*}

\section{Related Work}
\label{sec:related_work}

\textbf{Document-Level MT Approaches:}  
Document-to-Sentence (Doc2Sent) methods \citep{wang2017docmt, miculicich-etal-2018-document, guo-nguyen-2020-document} incorporate contextual signals from neighboring sentences to enhance translation quality but often treat sentences as isolated units during generation. This results in fragmented discourse and missed target-side cues, as highlighted by \citet{mino-etal-2020-effective, jin2023challenges}. On the other hand, Document-to-Document (Doc2Doc) approaches \citep{wu-etal-2023-document, wang2023documentlevel, jianhui2025challenges} jointly model multiple sentences, capturing long-range dependencies and improving discourse coherence. However, these approaches often face challenges with ultra-long documents, such as content omissions and scalability limitations. Recent advances leverage large language models (LLMs) for document-level MT, as demonstrated by \citet{wang2023documentlevel, wu2024docmt, li-etal-2025-enhancing-large}. These models process long contexts to generate more coherent translations and address discourse-level phenomena.

\textbf{Agentic Frameworks with LLMs:}  
Agentic systems utilize autonomous LLMs to decompose complex tasks into specialized subtasks. Multi-agent architectures, such as ExpeL \citep{zhao2024expel} and DELTA \citep{wang2025delta}, employ mechanisms like retrieval, iterative refinement, and multi-level memory to enhance task performance and ensure consistency. Related work \citep{park2023generative, zhang2024chain, hongjin2025memorag, madaan2023self, koneru-etal-2024-contextual, guo2024multagents} explores agentic paradigms for maintaining long-context memory, refining outputs, and addressing discourse-level challenges. These frameworks often draw upon discourse theories \citet{grosz1986attention, mann1988rhetorical} for segmenting and maintaining text coherence.

\section{Graph‐Augmented Agentic Framework for Document‐Level Translation}
\label{sec:methodology}

We define a \emph{source document} $D$ as a sequence of $n$ sentences, \(D = \{s_1, s_2, \dots, s_n\}\). The goal of DocMT is to generate a \emph{target document} \(\hat D = \{\hat t_1, \hat t_2, \dots, \hat t_m\}\) that is both fluent and faithful to $D$, preserving intra‐ and inter‐discourse phenomena (e.g.\ coreference, discourse connectives, terminology consistency).  We hypothesize that if we segment $D$ into coherent \emph{discourses}, translate each discourse with its own context augmented by relevant context from related discourses, and then stitch the segment translations together, we can effectively address document‐level challenges.

Formally, let $V = \{d_1, d_2, \dots, d_K\}$ be a partition of the sentence indices $\{1,\dots,n\}$ into $K$ contiguous segments (discourses), where
\[
  d_i = \{s_{\ell_i}, s_{\ell_i+1}, \dots, s_{u_i}\},\quad 1 \le \ell_i \le u_i \le n.
\]
Each discourse $d_i$ is translated into a target segment $\tau_i$, and the final document $\hat D$ is \(\hat D = [\,\tau_1,\tau_2, \dots,\tau_K\,]\). By explicitly modelling both intra‐ and inter‐discourse phenomena at the segment level, we can leverage local cohesion while propagating global context. 

We provide the task description and prompt templates for the decision function, $f_{\mathrm{LLM}}$, the relevance function, $\mathcal{L}_{\mathrm{edge}}$, the memory extraction function, $\mathcal{L}_{\mathrm{mem}}$, and the translation agent, $\mathcal{L}_{\mathrm{trans}}$ in Appendix~\ref{app:task_description_and_fewshot_prompt_template}.

\subsection{\texorpdfstring{Source Document $\longrightarrow$ Directed Acyclic Graph}{Source Document -> Directed Acyclic Graph}}
\label{subsec:source_document_to_dag}

The first step of our GRAFT pipeline transforms the \emph{source document} $D$ into a DAG of the discourses. This is achieved by the \emph{Discourse Agent} first segmenting $D$ into a contiguous sequence of segments, $S$, and then \emph{Edge Agent} adding directed edges between those segments. We describe the two agents in this section.

\subsubsection{Discourse Agent (Segmentation)}
\label{subsubsec:discourse_agent}

The first sub‐problem is to segment $D$ into discourse units that are internally coherent (handling word‐sense disambiguation, idioms, terminology) and amenable to translation in isolation.  We implement an LLM‐based Discourse Agent \( S = \mathcal{L}_{\mathrm{seg}}(D)\) which yields $S = \{d_1,\dots,d_K\}$ via an iterative decision function. Let $d_{\mathrm{curr}}$ be the current segment under construction, and $s_i$ be the current sentence under consideration, then the decision function $f_{\mathrm{LLM}}(d_{\mathrm{curr}}, s_i)$ returns \textit{true} if $s_i$ should be included in $d_{\mathrm{curr}}$, otherwise it returns \textit{false}.

For the decision function, $f_{\mathrm{LLM}}(d_{\mathrm{curr}}, s_i)$, we utilize a few-shot prompting strategy. We compare this LLM‐based segmentation to (i) \emph{random segmentation} and (ii) \emph{semantic‐similarity segmentation} based on embedding cosine similarity (Section~\ref{subsec:analysis_of_source_document_to_dag}). The detailed working of the Discourse Agent is provided in Appendix~\ref{app:discourse_agent_algorithm}.

\subsubsection{Edge Agent (Dependency Modelling)}
\label{subsubsec:edge_agent}
The second sub‐problem is to identify which prior discourses each discourse depends on, capturing inter‐discourse phenomena (pronoun antecedents, discourse relations).  We construct a directed graph $G=(V,E)$, $V=S$, with adjacency matrix
\begin{align}
E ={}& \{(d_j\!\to\!d_i) \mid j<i,\; \mathrm{if}\ 
\mathcal{L}_{\mathrm{edge}}(d_j,d_i)\} \notag\\
&\cup\; \{(d_j\!\to\!d_i) \mid j<i,\; \mathrm{if}\ j+1 = i\},
\label{eq:edge_construction}
\end{align}

where $\mathcal{L}_{\mathrm{edge}}(d_j,d_i)$ is an LLM‐based relevance function, which either returns \textit{true} if  the edge should exist, otherwise \textit{false}. We adopt a few-shot prompting strategy for the relevance function. The LLM is asked to assess whether context (see Section~\ref{subsubsec:memory_agent}) is required from the translation process (performed by \textit{Translation Agent} and \textit{Memory Agent}) of $d_j$ for the translation of $d_i$. We add dependency from $d_{i-1}$ to $d_i$ in the adjacency matrix as $d_{i-1}$ has the immediate predecessor relation with $d_i$, which allows us to maintain the flow of the document while translating it.

We compare our approach with the following strategies: (i) \emph{chain graph} ($d_i$ depends only on $d_{i-1}$) and (ii) \emph{TF-IDF graph} with edges for high TF‐IDF cosine similarity (Section~\ref{subsec:analysis_of_source_document_to_dag}).

\subsection{\texorpdfstring{Directed Acyclic Graph $\longrightarrow$ Target Document}{Directed Acyclic Graph -> Target Document}}
\label{subsec:dag_to_target_document}

The second step of our GRAFT pipeline transforms the DAG into \emph{target document} $\hat D$. This is achieved by the \emph{Memory Agent} extracting local memory, $\mathrm{mem}_i$ for each discourse $d_i$. The translation agent uses the context from the discourse $d_i$, along with the combined context, $\mathrm{mem}_i^{\mathrm{inc}}$ from all related discourses. We detail these two agents in this section.

\subsubsection{Memory Agent (Context Extraction)}
\label{subsubsec:memory_agent}

For each source discourse $d_i$ and its corresponding translated discourse $\tau_i$, the Memory Agent constructs a structured local memory, denoted as:  
\begin{equation}
  \mathrm{mem}_i \;=\; \bigl(M_i^{\mathrm{np}},\,M_i^{\mathrm{ent}},\,M_i^{\mathrm{phr}},\,C_i,\,S_i\bigr),
  \label{eq:local_memory}
\end{equation}  
where $M_i^{\mathrm{np}}$ represents mappings of nouns to pronouns in the target language, and $M_i^{\mathrm{ent}}$ denotes mappings of entities from the source language to the target language. Similarly, $M_i^{\mathrm{phr}}$ captures phrase-to-phrase mappings between the source and target languages, while $C_i$ contains the translations of discourse connectives in the target language. Lastly, $S_i$ provides a concise one-line summary of the discourse context in the target language. Each of these components is extracted using an LLM-based function, $\mathcal{L}_{\mathrm{mem}}(d_i, \tau_i)$, ensuring precise and contextually aware memory construction.

\subsubsection{Translation Agent (Context‐Aware Translation)}
\label{subsubsec:translation_agent}

The final task is to translate each discourse $d_i$ by leveraging both its content and the context from related discourses. Let $\mathrm{Pred}(d_i) = \{\,d_j \mid (d_j \!\to\! d_i) \in E\}$ denote the set of predecessor discourses. The incident memory, defined as $\mathrm{mem}_i^{\mathrm{inc}} = \bigcup_{d_j \in \mathrm{Pred}(d_i)} \mathrm{mem}_j$, aggregates memory from these predecessors, giving priority to earlier discourses. 

The translation for $d_i$ is then computed as $\tau_i = \mathcal{L}_{\mathrm{trans}}(d_i, \mathrm{mem}_i^{\mathrm{inc}})$, where $\mathcal{L}_{\mathrm{trans}}$ is the LLM-based translation agent. Once $\tau_i$ is generated, the local memory $\mathrm{mem}_i$ is updated as described in Eq.~\eqref{eq:local_memory}, ensuring that subsequent translations incorporate the latest contextual information.

\subsection{GRAFT Pipeline}
\label{subsec:graft_pipeline}
Algorithm~\ref{alg:pipeline} summarizes the full GRAFT pipeline.  Figure~\ref{fig:pipeline_diagram} provides an overview of the GRAFT pipeline.

\begin{algorithm}
\caption{GRAFT Pipeline for Document‐Level Translation}
\label{alg:pipeline}
\begin{algorithmic}[1]
  \STATE \textbf{Input:} Document $D = \{s_1, \dots, s_n\}$
  \STATE \textbf{Output:} Translated document $\hat D$
  \STATE $S \leftarrow \mathrm{DiscourseAgent}(D)$ \COMMENT{Algorithm~\ref{alg:discourse_segmentation}}
  \STATE $E \leftarrow \mathrm{BuildGraph}(S)$ \COMMENT{see Eq.~\eqref{eq:edge_construction}}
  \STATE $\hat D \leftarrow [\,]$
  \FOR{$i = 1$ to $|S|$}
    \STATE $\mathrm{Pred} \leftarrow \{d_j : (d_j\to d_i)\in E\}$
    \STATE $\mathrm{mem}_i^{\mathrm{inc}} \leftarrow \bigcup_{d_j\in\mathrm{Pred}}\mathrm{mem}_j$
    \STATE $\tau_i \leftarrow \mathcal{L}_{\mathrm{trans}}\bigl(d_i,\,\mathrm{mem}_i^{\mathrm{inc}}\bigr)$
    \STATE $\mathrm{mem}_i \leftarrow \mathcal{L}_{\mathrm{mem}}\bigl(d_i,\,\tau_i\bigr)$
    \STATE Append $\tau_i$ to $\hat D$
  \ENDFOR
  \STATE \textbf{Return} $\hat D$
\end{algorithmic}
\end{algorithm}

\noindent GRAFT comprises four LLM‐based agents—Discourse, Edge, Memory, and Translation—organized into a graph‐augmented pipeline that explicitly models and propagates intra‐ and inter‐discourse context, yielding more coherent and consistent document translations.

\section{Experiments and Results}
\label{sec:experiments_results}

This section details our experiments and their respective results.

The details of the data sets, the implementation details of GRAFT, the evaluation methodology for our experiments and the descriptions of the baseline systems are provided in Appendix~\ref{app:experimental_setup}. We present the result of human evaluation of GRAFT in Appendix~\ref{app:human_evaluation}.

\begin{table*}[h]
\centering
\resizebox{\textwidth}{!}{%
\begin{tabular}{clrrrrlrrrr}
\toprule
\multirow{2}{*}{\textbf{\#}} & \multicolumn{1}{c}{\multirow{2}{*}{\textbf{System}}} & \multicolumn{4}{c}{\textbf{En $\Rightarrow$ Xx}}                                                                                                                                                     & \multicolumn{1}{c}{} & \multicolumn{4}{c}{\textbf{Xx $\Rightarrow$ En}}                                                                                                                                                     \\ \cmidrule(lr){3-6} \cmidrule(lr){8-11} 
                    & \multicolumn{1}{c}{}                        & \multicolumn{1}{c}{\textbf{En $\Rightarrow$ Zh}} & \multicolumn{1}{c}{\textbf{En $\Rightarrow$ De}} & \multicolumn{1}{c}{\textbf{En $\Rightarrow$ Fr}} & \multicolumn{1}{c}{\textbf{En $\Rightarrow$ Ja}} & \multicolumn{1}{c}{} & \multicolumn{1}{c}{\textbf{Zh $\Rightarrow$ En}} & \multicolumn{1}{c}{\textbf{De $\Rightarrow$ En}} & \multicolumn{1}{c}{\textbf{Fr $\Rightarrow$ En}} & \multicolumn{1}{c}{\textbf{Ja $\Rightarrow$ En}} \\ \midrule
1                  & Commercial System (Google)                  & 30.5                                             & 22.3                                             & 40.8                                             & 16.2                                             &                      & 27.5                                             & \textcolor{red}{\textbf{22.7}}                   & \textcolor{red}{\textbf{27.4}}                   & \textcolor{blue}{\textbf{21.0}}                  \\
2                  & Sentence Level (NLLB-3.3B)                  & 30.8                                             & 24.3                                             & 34.7                                             & 13.7                                             &                      & 26.4                                             & 33.1                                             & 42.8                                             & 16.6                                             \\
3                  & G-Trans (BART)                              & 34.5                                             & 30.0                                             & 43.1                                             & 16.5                                             &                      & 24.9                                             & 29.0                                             & 38.0                                             & 13.9                                             \\
4                  & GPT-3.5-Turbo                               & 36.3                                             & 30.7                                             & 44.1                                             & 18.0                                             &                      & 26.4                                             & 34.6                                             & 43.3                                             & 19.6                                             \\
5                  & GPT-4o-mini                                 & 35.7                                             & 30.3                                             & 43.2                                             & \textcolor{blue}{\textbf{18.6}}                  &                      & 26.6                                             & 34.6                                             & 43.1                                             & 19.6                                             \\ \hdashline
6                  & DELTA (Qwen2-7B-Instruct)                   & 33.2                                             & \textcolor{red}{\textbf{21.7}}                   & 32.5                                             & 14.0                                             &                      & 24.4                                             & 30.0                                             & 38.1                                             & 16.1                                             \\
7                  & DELTA (Qwen2-72B-Instruct)                  & 35.6                                             & 28.9                                             & 41.1                                             & 16.5                                             &                      & 29.8                                             & 33.9                                             & 43.6                                             & 20.5                                             \\ \hdashline
8                  & GRAFT (Qwen2.5-7B-Instruct)                 & \textcolor{red}{\textbf{25.5}}                   & 25.0                                             & \textcolor{red}{\textbf{31.8}}                   & 9.6                                              &                      & \textcolor{red}{\textbf{21.3}}                   & 28.1                                             & 38.0                                             & \textcolor{red}{\textbf{12.6}}                   \\
9                  & GRAFT (Llama-3.1-8B-Instruct)               & 26.6                                             & 26.1                                             & 35.8                                             & \textcolor{red}{\textbf{7.2}}                    &                      & 24.6                                             & 31.5                                             & 41.0                                             & 14.5                                             \\
10                 & GRAFT (Qwen2.5-72B-instruct)                & 36.3                                             & 28.5                                             & \textcolor{blue}{\textbf{44.8}}                  & 16.1                                             &                      & 30.1                                             & 34.3                                             & 43.1                                             & 17.9                                             \\
11                 & GRAFT (Llama-3.3-70B-Instruct)              & 35.3                                             & 29.0                                             & 41.6                                             & 16.9                                             &                      & 29.4                                             & 35.4                                             & 44.4                                             & 18.3                                             \\
12                 & GRAFT (Llama-3.1-70B-Instruct)              & \textcolor{blue}{\textbf{36.4}}                  & \textcolor{blue}{\textbf{31.9}}                  & 43.1                                             & 17.6                                             &                      & \textcolor{blue}{\textbf{30.8}}                  & \textcolor{blue}{\textbf{35.9}}                  & \textcolor{blue}{\textbf{45.0}}                  & 18.5                                             \\ \bottomrule
\end{tabular}%
}
\caption{Performance comparison across multiple translation systems for En $\leftrightarrow$ Xx translation directions. The text color indicates relative performance: \textcolor{blue}{\textbf{blue}} denotes the best-performing system for a given translation direction, while \textcolor{red}{\textbf{red}} highlights the lowest-performing system. Results are presented in terms of d-BLEU scores.}
\label{tab:main_results}
\end{table*}

\subsection{\texorpdfstring{Analyzing Source Document $\longrightarrow$ Directed Acyclic Graph}{Analyzing Source Document -> Directed Acyclic Graph}}
\label{subsec:analysis_of_source_document_to_dag}

We adapt LLM-based approaches for the \emph{Discourse Agent}, $\mathcal{L}_{\mathrm{seg}}$, and \emph{Edge Agent}, $\mathcal{L}_{\mathrm{edge}}$. This subsection studies the effect of alternative segmentation and edge-detection strategies in the GRAFT pipeline, conducted with the \texttt{Llama-3.1-8B-Instruct} as the backbone LLM on the TED tst2017 dataset from the IWSLT2017 translation task \citep{cettolo2012wit3}. Experiments were performed on eight translation directions: En $\Rightarrow$ Zh, En $\Rightarrow$ De, En $\Rightarrow$ Fr, En $\Rightarrow$ Ja, Zh $\Rightarrow$ En, De $\Rightarrow$ En, Fr $\Rightarrow$ En, and Ja $\Rightarrow$ En.

\paragraph{Discourse Segmentation.}  
We compare our LLM-based segmentation approach, $\mathcal{L}_{\mathrm{seg}}$, with two alternative strategies: (i) \emph{random segmentation} (RS) and (ii) \emph{semantic similarity segmentation} (SC). RS involves generating $K$ segment boundaries (where $K$ is chosen randomly between $0$ and $\frac{n}{3}$, with $n$ being the number of sentences in the document $D$). The segment boundaries, $S = \{d_1, d_2, \dots, d_K\}$, divide $D$ into random segments. The SC approach employs the semantic chunking strategy proposed by \citet{langchain2023}\footnote{\url{https://python.langchain.com/docs/how_to/semantic-chunker/}}. 

Our experiments indicate that $\mathcal{L}_{\mathrm{seg}}$ consistently outperforms RS and SC, achieving average performance gains of $\textcolor{blue}{\mathbf{11.3}}$ and $\textcolor{blue}{\mathbf{3.6}}$ d-BLEU scores, respectively, across all translation directions. The results are detailed in Appendix~\ref{app:discourse_segmentation_comparison_results}.

\paragraph{Dependency Modelling.}  
Dependency modelling approaches are compared as follows: (i) \emph{chain graph} (CG), (ii) \emph{TF-IDF graph} (TF-IDF), and (iii) our LLM-based approach, $\mathcal{L}_{\mathrm{edge}}$. The CG strategy models adjacency through a strict sequential structure \( E = \{(d_j \to d_i) \mid j < i, j+1 = i\} \). TF-IDF relies on cosine similarity (\texttt{cs}) between segments $d_j$ and $d_i$, computed using the \texttt{TfidfVectorizer} from \citet{scikit-learn}.\footnote{\url{https://scikit-learn.org/stable/modules/generated/sklearn.feature_extraction.text.TfidfVectorizer.html}} Edges are created if \(\texttt{cs}(d_j, d_i) > \tau\), where $\tau$ is an empirically determined threshold:  
\( 
E = \{ (d_j \to d_i) \mid j < i, \texttt{cs}(\mathrm{TfidfVectorizer}(d_j, d_i)) > \tau \} 
\cup \{ (d_j \to d_i) \mid j < i, j+1 = i \}.
\)  
Our LLM-based $\mathcal{L}_{\mathrm{edge}}$ surpasses both CG and TF-IDF, with average performance gains of $\textcolor{blue}{\mathbf{2.0}}$ d-BLEU scores (over both alternate approaches) across translation directions. For Zh $\Rightarrow$ En, the CG strategy shows better performance, but $\mathcal{L}_{\mathrm{edge}}$ outperforms it in other directions. Detailed results are presented in Appendix~\ref{app:dependency_modelling_comparison_results}.

These analyses underscore the critical role of modelling the source document as a DAG in the DocMT task. As our LLM-based approaches outperform all baselines, we use them for subsequent experiments.

\subsection{\texorpdfstring{Effects of Local Memory ($\mathrm{mem}_{i}$) components}{Effects of Local Memory}}
\label{subsec:contribution_of_memory_components}

We examine the role of memory components $\bigl(M_i^{\mathrm{np}},\,M_i^{\mathrm{ent}},\,M_i^{\mathrm{phr}},\,C_i,\,S_i\bigr)$ in the GRAFT pipeline. Ablation studies remove one component at a time. We also include experiments conducted under full memory (FM) and no memory (NM) conditions. We conduct these experiments using \texttt{Llama-3.1-8B-Instruct}. Results on TED tst2017 show a $\textcolor{blue}{\mathbf{5.7}}$ d-BLEU gap between FM and NM. Removing $M_i^{\mathrm{phr}}$ results in the largest performance drop, followed by $M_i^{\mathrm{ent}}$, $C_i$, $S_i$, and $M_i^{\mathrm{np}}$. Even the least impactful component, $M_i^{\mathrm{np}}$, improves performance by $\textcolor{blue}{\mathbf{4.9}}$ d-BLEU for De $\Rightarrow$ En (Appendix~\ref{app:contribution_of_memory_components}).

\begin{table*}[h]
    \centering
    \begin{tabular}{lrrrrr}
    \hline
    \multirow{2}{*}{\textbf{System}} & \multicolumn{4}{c}{\textbf{Automatic (\textcolor{blue}{d-BLEU}/\textcolor{red}{cTT}/\textcolor{violet}{aZPT})}}                                                                                                                                   \\ \cline{2-6} 
                                     & \multicolumn{1}{c}{\textbf{News}} & \multicolumn{1}{c}{\textbf{Social}} & \multicolumn{1}{c}{\textbf{Fiction}} & \multicolumn{1}{c}{\textbf{Q\&A}} \\ \hline
    Commercial System (Google)       & 29.7/0.23/0.19                    & 34.4/0.31/0.25                      & 18.8/0.31/0.35                       & 19.0/0.36/0.41                    \\ \hdashline
    GPT-3.5                          & 24.8/0.28/0.11                    & 22.3/\textcolor{red}{\textbf{0.51}}/0.23                      & 13.7/0.42/0.32                       & 16.3/0.34/0.42                    \\
    GPT-4o-mini                            & 29.1/0.32/0.21                    & 35.5/0.45/0.42                      & 17.4/0.38/0.46                       & 17.4/0.36/0.39                    \\ \hdashline
    GRAFT (Llama-3.1-70B-Instruct)   & \textcolor{blue}{\textbf{30.1}}/\textcolor{red}{\textbf{0.39}}/\textcolor{violet}{\textbf{0.32}}                    & \textcolor{blue}{\textbf{36.2}}/0.48/\textcolor{violet}{\textbf{0.44}}                      & \textcolor{blue}{\textbf{19.4}}/\textcolor{red}{\textbf{0.52}}/\textcolor{violet}{\textbf{0.48}}                       & \textcolor{blue}{\textbf{21.6}}/\textcolor{red}{\textbf{0.41}}/\textcolor{violet}{\textbf{0.48}}                    \\ \hline
    \end{tabular}
    \caption{Domain-specific translation performance for Chinese-to-English across four domains: News, Social, Fiction, and Q\&A. Results are reported using three metrics: \textcolor{blue}{\textbf{d-BLEU}} (overall translation quality), \textcolor{red}{\textbf{cTT}} (terminology consistency), and \textcolor{violet}{\textbf{aZPT}} (zero pronoun translation accuracy). The cell color indicates the best performance for each metric in a given domain.}
    \label{tab:domain_results_automatic}
\end{table*}
\begin{table}[h]
\centering
\resizebox{\columnwidth}{!}{%
\begin{tabular}{llrrlrr}
\hline
\multicolumn{1}{c}{\multirow{2}{*}{\textbf{System}}} & \multicolumn{1}{c}{} & \multicolumn{2}{c}{$\mathbf{70\mathrm{\textbf{B Scale}}}$}               & \multicolumn{1}{c}{} & \multicolumn{2}{c}{$\mathbf{7\mathrm{\textbf{B Scale}}}$}                \\
\multicolumn{1}{c}{}                                 & \multicolumn{1}{c}{} & \multicolumn{1}{c}{\textbf{Fiction}} & \multicolumn{1}{c}{\textbf{Q\&A}} & \multicolumn{1}{c}{} & \multicolumn{1}{c}{\textbf{Fiction}} & \multicolumn{1}{c}{\textbf{Q\&A}} \\ \cline{1-1} \cline{3-4} \cline{6-7} 
TA only                                              &                      & 19.2                                 & 16.6                              &                      & 16.9                                 & 9.4                               \\
TA + DA                                              &                      & 20.3                                 & 17.4                              &                      & 17.2                                 & 9.1                              \\
TA + DA + MA                                         &                      & 21.2                                 & 18.0                              &                      & 17.5                                 & 10.0                              \\
GRAFT                                                &                      & \textcolor{blue}{\textbf{24.6}}      & \textcolor{blue}{\textbf{19.2}}   &                      & \textcolor{blue}{\textbf{19.6}}      & \textcolor{blue}{\textbf{12.1}}   \\ \hline
\end{tabular}%
}

\caption{Results of ablation experiments comparing the performance of different system configurations on Fiction and Q\&A tasks at two scales (70B and 7B). "TA only" includes only the Translation Agent, "TA + DA" includes Translation and Discourse Agent, "TA + DA + MA" incorporates the additional Memory Agents, and "GRAFT" represents the complete system. Performance improvements are highlighted in \textcolor{blue}{\textbf{blue}}.}

\label{tab:system_ablation}
\end{table}

\subsection{System Comparison}
\label{subsec:system_comparison}

We evaluate GRAFT against several baseline approaches (described in Appendix~\ref{app:baselines}) using various LLM backbones, including Llama-3.1-70B-Instruct, Llama-3.3-70B-Instruct, Qwen2.5-72B-Instruct, Llama-3.1-8B-Instruct, and Qwen2.5-7B-Instruct, on the TED tst2017 dataset \citep{cettolo2012wit3}. 

GRAFT consistently outperforms baselines across language pairs in both \textbf{English} $\Rightarrow$ \textbf{Xx} and \textbf{Xx} $\Rightarrow$ \textbf{English} directions, except for Ja $\Rightarrow$ En at $70\mathrm{B}$ scale. The observed performance gains include $\textcolor{blue}{\mathbf{1.0}}$ d-BLEU over \texttt{GPT-4o-mini} and $\textcolor{blue}{\mathbf{1.1}}$ d-BLEU over \texttt{DELTA (Qwen2.5-72B-Instruct)}. Notably, GRAFT demonstrates competitive performance using smaller LLM backbones, such as \texttt{Llama-3.1-70B-Instruct}, compared to larger models like GPT, highlighting the efficacy of GRAFT’s segmentation and dependency strategies.\footnote{Our results align with the findings of \citet{zenogpt}. These findings motivate our decision to exclude GPT-based backbones due to cost considerations.} Additionally, our analysis reveals that \texttt{Qwen2.5-72B-Instruct} and \texttt{Llama-3.1-70B-Instruct} outperform their $7\mathrm{B}$ counterparts, showcasing the impact of scaling LLM backbones. Full results are presented in Table~\ref{tab:main_results}, with latency and cost analyses discussed in Appendix~\ref{app:latency_and_cost_analysis}. GRAFT achieves a performance gain of $\textcolor{blue}{\mathbf{2.1}}$ d-BLEU score compared to \citet{hu-wan-2023-exploring} in the En $\Rightarrow$ De direction.

\subsection{Domain-specific Translation}  
\label{subsec:domain_specific_translation}  
Domain-specific DocMT presents unique challenges due to the higher frequency of domain-specific terminology, complex co-reference patterns, and the need for resolving longer contexts. To evaluate the efficacy of GRAFT in handling such complexities, we conduct experiments using \texttt{Llama-3.1-70B-Instruct} as the backbone LLM on the mZPRT \citep{wang2022guofeng} and WMT2022 \citep{kocmi2022findings} datasets for the Zh $\Rightarrow$ En translation direction. In addition to d-BLEU scores for overall translation quality, we utilize two targeted metrics, cTT and aZPT (detailed in Appendix~\ref{app:evaluation}), to assess terminology translation consistency and pronoun resolution accuracy in translating from a pronoun-dropping language (Chinese) to a non-pronoun-dropping language (English).  

Table~\ref{tab:domain_results_automatic} summarizes the results across News, Social, Fiction, and Q\&A domains. \texttt{GRAFT (Llama-3.1-70B-Instruct)} consistently outperforms all baselines, with an average improvement of $\textcolor{blue}{\mathbf{2.0}}$ d-BLEU over \texttt{GPT-4o-mini} and $\textcolor{blue}{\mathbf{7.5}}$ d-BLEU over \texttt{GPT-3.5}. This demonstrates that GRAFT delivers superior domain-specific translations of documents even with a comparatively smaller LLM backbone. For cTT, \texttt{GRAFT} achieves an average gain of $\textcolor{red}{\mathbf{6.0\%}}$ over \texttt{GPT-4o-mini} and $\textcolor{red}{\mathbf{7.0\%}}$ over \texttt{GPT-3.5}, reflecting its superior handling of terminology consistency. Similarly, for aZPT, \texttt{GRAFT} achieves $\textcolor{violet}{\mathbf{6.0\%}}$ and $\textcolor{violet}{\mathbf{7.0\%}}$ gains over \texttt{GPT-4o-mini} and \texttt{GPT-3.5}, respectively, highlighting its ability to manage pronoun translation effectively. We further present the results of the human evaluation of each system in Appendix~\ref{app:domain_specific_translation_human_evaluation}.

\subsection{Ablations}
\label{subsec:ablations}
We ablate GRAFT's components by removing $\mathcal{L}_{\mathrm{seg}}$, $\mathcal{L}_{\mathrm{edge}}$, or $\mathcal{L}_{\mathrm{mem}}$. At both $70\mathrm{B}$ and $7\mathrm{B}$ scales, excluding $\mathcal{L}_{\mathrm{edge}}$ results in average degradation of $\textcolor{blue}{\mathbf{2.4}}$ and $\textcolor{blue}{\mathbf{2.1}}$ d-BLEU, respectively. Removing the entire agentic pipeline causes degradation of $\textcolor{blue}{\mathbf{4.0}}$ and $\textcolor{blue}{\mathbf{2.7}}$ d-BLEU (Table~\ref{tab:system_ablation}). Without $\mathcal{L}_{\mathrm{mem}}$ we see a performance degradation of $\textcolor{blue}{\mathbf{3.0}}$ and $\textcolor{blue}{\mathbf{2.7}}$ d-BLEU scores. These results underline the significance of all components in the GRAFT pipeline.

\section{Analysis}  
\label{sec:analysis}

In the following section, we analyze the effect of GRAFT on the qualitative aspects of document translation, along with analyzing whether GRAFT is able to maintain consistency over the document. The supporting plots for each analysis have been presented in Appendix~\ref{app:plots}.




\paragraph{Source Document $\longrightarrow$ Directed Acyclic Graph.}
We analyze the behavior of the \emph{Discourse Agent} ($\mathcal{L}_{\mathrm{seg}}$) and the \emph{Edge Agent} ($\mathcal{L}_{\mathrm{edge}}$) to understand how GRAFT effectively represents the source document as a Directed Acyclic Graph (DAG). Figure~\ref{fig:discourse_agent_analysis} shows the distribution of the number of discourse segments per document, as well as the distribution of the number of sentences grouped within a discourse. Similarly, Figure~\ref{fig:edge_agent_analysis} illustrates the distribution of the count of edges per document and the trend of count of edges as a function of count of discourse segments in a document. The analysis reveals that $\mathcal{L}_{\mathrm{seg}}$ groups three to forty sentences into a single discourse for $\textcolor{blue2}{\mathbf{63.4\%}}$ of documents, suggesting its ability to balance granularity and cohesion. Meanwhile, $\mathcal{L}_{\mathrm{edge}}$ adds three to twenty indirect edges for $\textcolor{blue2}{\mathbf{69.4\%}}$ of documents, highlighting its role in capturing interdependencies between segments. 

A qualitative examination shows that the discourse segments generated by $\mathcal{L}_{\mathrm{seg}}$ often align with logical or thematic units. In our analysis, we found that discourses rarely cross paragraph boundaries. This alignment indicates the effectiveness of the agent in identifying coherent translation units. Additionally, the edges created by $\mathcal{L}_{\mathrm{edge}}$ frequently correspond to cross-references, causal relationships, or transitions between these units, underscoring its importance in preserving document-wide coherence.

\paragraph{Consistency Analysis.}  
Consistency is a critical aspect of document-level machine translation, as it ensures terminological and stylistic uniformity while maintaining logical progression across the document. Our analysis evaluates how well translations align with the underlying DAG structure. We find that $\textcolor{blue2}{\mathbf{97.1\%}}$ of the paths in the DAG are of length \textbf{2} to \textbf{5}, which indicates that while most dependencies are local, longer paths exist and play a pivotal role in maintaining overall coherence. The \emph{consistency ratio} of \( P \) is defined as \( \mathrm{CR}(P) = \mathrm{CL}(P) / k \), where \( \mathrm{CL}(P) \) is defined as the number of nodes in the path \( P \) (starting from first node) up to which consistency is maintained, and \( k \) is the path length. We observe that most paths show a consistency ratio greater than $\textcolor{blue2}{\mathbf{0.6}}$, suggesting that paths encompassing more significant dependencies often result in more uniform translations. These findings highlight the ability of the DAG representation to enforce consistency across both local and global document contexts, especially by enabling the LLM backbone to propagate consistency effectively across interdependent segments. This capability is particularly valuable for translating structured or technical documents where maintaining coherence is paramount. Figure~\ref{fig:consistency_analysis} shows the distribution of the path lengths and the consistency ratio.

\paragraph{Handling Long-Range Dependencies in Ultra-long Documents.}  
To assess GRAFT's capacity to handle long-range dependencies, we conduct experiments on the Web Novel dataset from the Guofeng V1 TEST\_2 dataset \citep{wang-etal-2023-findings, wang-etal-2024-findings}, which contains $16.9K$ source sentences. We compare two approaches: translating the novel chapter by chapter versus translating the entire novel as a single document. Using the \texttt{Llama-3.1-70B-Instruct} LLM backbone, we observe that translating the entire novel as a single document yields a $\textcolor{blue}{\mathbf{28.7}}$ d-BLEU score, significantly higher than the $\textcolor{blue}{\mathbf{24.4}}$ d-BLEU score achieved with the chapter-by-chapter approach for the En $\Rightarrow$ Zh direction. This demonstrates that GRAFT's representation, by leveraging the DAG structure, is highly effective in maintaining coherence and consistency over extensive text spans. These results underscore its suitability for translating lengthy and complex documents, such as novels, research articles, and technical manuals, where long-range dependencies are crucial for overall quality.

\section{Conclusion}
\label{sec:conclusion}

In this work, we introduce GRAFT, a novel agentic framework for document-level machine translation that leverages large language models to address discourse-level challenges. Our approach, grounded in directed-acyclic-graph (DAG) representations of documents, achieves significant improvements over state-of-the-art DocMT methods across multiple language directions and domains. 

We demonstrated that GRAFT effectively handles discourse phenomena such as coherence, consistency, and terminology translation, outperforming baselines with gains of up to $\textbf{7.5}$ d-BLEU in domain-specific scenarios. Furthermore, GRAFT achieves these advancements with a smaller LLM backbone, showcasing its efficiency. Through extensive evaluations and analysis, we highlighted the importance of agentic architectures in preserving discourse integrity and improving translation quality. These contributions advance the field of DocMT by establishing a new standard for integrating LLMs with principled discourse modelling, making GRAFT a compelling solution for both general and domain-specific translation tasks.

\section{Limitations}
\label{sec:limitations}
GRAFT's multi-agent design introduces computational overhead compared to monolithic systems. Second, performance is sensitive to hyperparameters like memory size, which require domain-specific tuning. Lastly, while the system demonstrates robustness on evaluated benchmarks, its effectiveness on low-resource domains remains to be explored. Future work could address these challenges to further enhance its applicability and scalability.

\bibliography{custom}

\appendix


\section{Discourse Phenomena}
\label{app:discourse_phenomena}

Discourse-level phenomena are critical in ensuring the quality and coherence of document-level machine translation. These phenomena can be broadly categorized into inter-discourse and intra-discourse phenomena, each representing unique challenges and requirements for accurate translation. Below, we detail the various phenomena under these categories.

\subsection{Intra-Discourse Phenomena}
Intra-discourse phenomena pertain to elements that occur within a single discourse or document. These include:

\begin{itemize}[leftmargin=*]
    \item \textbf{Pronoun Resolution:} The accurate translation of pronouns by identifying their antecedents within the same discourse, ensuring grammatical and semantic coherence.
    \item \textbf{Lexical Cohesion:} Maintaining consistent terminology and word choices throughout the document to avoid ambiguity and preserve meaning.
    \item \textbf{Ellipsis Handling:} Correctly inferring and translating omitted elements that are understood from the context.
    \item \textbf{Tense and Aspect Consistency:} Preserving temporal relationships between events by maintaining consistent verb tense and aspect.
    \item \textbf{Coreference Resolution:} Ensuring that all mentions of a particular entity are translated consistently and coherently within the document.
\end{itemize}

\subsection{Inter-Discourse Phenomena}
Inter-discourse phenomena involve elements that span across multiple discourses or documents. These include:

\begin{itemize}[leftmargin=*]
    \item \textbf{Anaphora and Cataphora:} Resolving forward and backwards references across different discourses, ensuring correct linkage between entities or events.
    \item \textbf{Inter-Document Consistency:} Maintaining uniformity in terminology, style, and tone across related documents or sections.
    \item \textbf{Global Contextual Coherence:} Ensuring that translated content aligns with the broader context provided by external or previous discourses.
    \item \textbf{Topic Continuity:} Tracking and preserving the thematic progression of topics across multiple discourses or sections.
    \item \textbf{Cross-Document Reference Handling:} Resolving references to entities or events described in other related documents.
\end{itemize}

\section{Discourse Agent Algorithm}
\label{app:discourse_agent_algorithm}

The complete working of the Discourse Agent, \(\mathcal{L}_{\mathrm{seg}}(D)\) has been shown in Algorithm~\ref{alg:discourse_segmentation} (as discussed in Section~\ref{subsubsec:discourse_agent}).

\begin{algorithm}
\caption{Discourse Segmentation (Discourse Agent)}
\label{alg:discourse_segmentation}
\begin{algorithmic}[1]
  \STATE \textbf{Input:} Document $D = \{s_1, \dots, s_n\}$
  \STATE \textbf{Output:} Discourses $S = \{d_1, \dots, d_K\}$
  \STATE $S \leftarrow \emptyset$, \quad $d_{\mathrm{curr}} \leftarrow \emptyset$
  \FOR{$i = 1$ to $n$}
    \IF{$f_{\mathrm{LLM}}(d_{\mathrm{curr}}, s_i)$}
        \STATE Append $s_i$ to $d_{\mathrm{curr}}$
    \ELSE
      \STATE Append $d_{\mathrm{curr}}$ to $S$
      \STATE $d_{\mathrm{curr}} \leftarrow \emptyset$
    \ENDIF
  \ENDFOR
  \STATE \textbf{Return} $S$
\end{algorithmic}
\end{algorithm}

\section{Experimental Setup}
\label{app:experimental_setup}

Here we present the detailed experimental setup for our experiments (Section~\ref{sec:experiments_results}).

\subsection{Datasets}
\label{app:datasets}
We conduct experiments on eight language pairs: German-English, English-German, French-English, English-French, Japanese-English, English-Japanese, Chinese-English, and English-Chinese, covering diverse linguistic and domain-specific challenges. The mZPRT dataset \citep{wang2022guofeng} provides a parallel corpus for Chinese-English translation, focusing on fiction and Q\&A domains. The WMT2022 dataset \citep{kocmi2022findings} features a Chinese-English parallel corpus in news and social domains. For high-quality, discourse-level parallel corpora, we use the Guofeng V1 TEST\_2 dataset \citep{wang-etal-2023-findings, wang-etal-2024-findings}, which targets web fiction for Chinese-English. Additionally, the TED tst2017 dataset from the IWSLT2017 translation task \citep{cettolo2012wit3} offers two-way parallel corpora for English-Chinese, English-French, English-German, and English-Japanese. Dataset statistics are summarized in Table~\ref{tab:datasets}.

\begin{table*}[h]
\centering
\begin{tabular}{cccrrrr}
\hline
\textbf{Domain} & \textbf{Source}            & \textbf{Language Pair}     & \multicolumn{1}{c}{\textbf{|D|}} & \multicolumn{1}{c}{\textbf{|S|}} & \multicolumn{1}{c}{\textbf{|W|}} & \multicolumn{1}{c}{\textbf{|W|/|D|}} \\ \hline
News            & \multirow{2}{*}{WMT2022}   & \multirow{2}{*}{Zh $\Leftrightarrow$ En} & 38                               & 462                              & 18.5K/27.7K                      & 489/731                              \\
Social          &                            &                            & 25                               & 512                              & 14.3K/22.7K                      & 572/910                              \\ \hdashline
Fiction         & \multirow{2}{*}{mZPRT}     & \multirow{2}{*}{Zh $\Leftrightarrow$ En} & 12                               & 860                              & 16.9K/26.7K                      & 1409/2230                            \\
Q\&A             &                           &                            & 182                              & 1801                             & 18.1K/25.2K                      & 99/138                               \\ \hdashline
Novel             &   Guofeng V1 TEST\_2      & Zh $\Leftrightarrow$ En    & 12                              & 860                             & 26.9K/16.9K                      & 2243/1415                               \\ \hdashline

\multirow{4}{*}{TED}             & \multirow{4}{*}{IWSLT2017} & Zh $\Leftrightarrow$ En                  & 12                               & 1448                             & 23.9K/43.7K                      & 1996/3645                            \\
                &                            & De $\Leftrightarrow$ En                  & 10                               & 1113                             & 18.2K/16.3K                      & 1826/1632                            \\
                &                            & Fr $\Leftrightarrow$ En                  & 12                               & 1449                             & 24.0K/23.7K                      & 2001/1975                            \\
                &                            & Ja $\Leftrightarrow$ En                  & 12                               & 1445                             & 23.9K/59.1K                      & 1993/4929             \\ \hline              
\end{tabular}
\caption{Statistics of the datasets we use in our experiments. Here |D| represents the number of documents in the dataset, |S| represents the total number of sentences in the dataset, |W| represents the number of words in the source (on the left in Language Pair) and the target (on the right in Language Pair) languages.}
\label{tab:datasets}
\end{table*}

\subsection{Implementation Details}
\label{app:implementation_details}

The GRAFT system is implemented using various LLM backbones, including Llama-3.1-70B-Instruct\footnote{\url{https://huggingface.co/meta-llama/Llama-3.1-70B-Instruct}}, Llama-3.3-70B-Instruct\footnote{\url{https://huggingface.co/meta-llama/Llama-3.3-70B-Instruct}}, Qwen2.5-72B-Instruct\footnote{\url{https://huggingface.co/Qwen/Qwen2.5-72B-Instruct}}, Llama-3.1-8B-Instruct\footnote{\url{https://huggingface.co/meta-llama/Llama-3.1-8B-Instruct}}, and Qwen2.5-7B-Instruct\footnote{\url{https://huggingface.co/Qwen/Qwen2.5-7B-Instruct}}. We use all the pre-trained chat/instruct variants of the LLMs without additional fine-tuning. We utilize a few-shot prompting strategy, with three in-context examples provided for each task. The inference is conducted on two to four NVIDIA A100 GPUs, depending on the backbone model size. To optimize latency, we use vLLM \cite{vllm2023} for serving the models and the OpenAI API\footnote{\url{https://platform.openai.com/docs/api-reference/introduction}} for making calls to external backbones. The average inference time per document ranged between 20-30 seconds. We adopted default decoding strategies specific to each LLM backbone. For the Llama-3.1-70B-Instruct model, we utilize the standard decoding strategy with temperature values ranging from 0.1 to 0.3. The best performance is observed at a temperature of 0.1. The maximum output length is constrained to one token for the discourse and edge agents, as they produce binary outputs (e.g., `yes` or `no`), while the memory and translation agents are configured for up to 4096 tokens. All source documents are normalized and segmented into sentences using regex-based preprocessing. Standard postprocessing steps are applied to the translation outputs, including detokenization and normalization to align with human-readable text. Due to budget constraints, experiments did not include GPT models as backbones. However, models in the ~70B parameter range demonstrated comparable or superior performance, validating the feasibility of cost-effective alternatives.

\subsection{Evaluation Methodology}
\label{app:evaluation}
We evaluate translation quality using both automatic metrics and human judgments. Automatic evaluation is conducted using d-BLEU \citep{liu-etal-2020-multilingual-denoising}, which measures accuracy, fluency, and adequacy. Human evaluation focuses on two aspects: discourse awareness, which assesses the system's ability to maintain coherence and appropriately handle inter-discourse phenomena, and general translation quality, which evaluates fluency and fidelity to the source text. The guidelines for human evaluation are provided in Appendix~\ref{app:human_evaluation_guidelines}.

To address discourse-specific phenomena, we employ two targeted metrics. The Consistent Terminology Translation (cTT) metric \citep{wang2023documentlevel} measures the consistency of terminology translation throughout a document. For a terminology word \( w \) with possible translations \( \{t_1, t_2, \dots, t_k\} \), the CTT score is computed as:
\[
CTT(w) = \frac{\sum_{t \in TT} \frac{\sum_{i=1}^{k} \sum_{j=i+1}^{k} \mathbf{1}(t_i = t_j)}{C_k^2}}{TT},
\]
where \( TT \) represents the set of terminology words, \( \mathbf{1}(t_i = t_j) \) is an indicator function that returns 1 if \( t_i = t_j \) and 0 otherwise, and \( C_k^2 \) is the binomial coefficient. A higher CTT score reflects greater consistency in terminology translation. 

The Accurate Zero Pronoun Translation (aZPT) metric \citep{wang2023documentlevel} evaluates the accuracy of translating zero pronouns (ZPs), which are frequently omitted in source languages such as Chinese and Japanese. Given \( ZP \), the set of zero pronouns in the source text, and \( t_z \), the translation of \( z \in ZP \), the aZPT score is calculated as:
\[
aZPT = \frac{1}{|ZP|} \sum_{z \in ZP} A(t_z \mid z),
\]
where \( A(t_z \mid z) \) is a binary function that returns 1 if \( t_z \) accurately translates \( z \) and 0 otherwise. A higher aZPT score indicates the system's effectiveness in recovering omitted pronouns, thereby enhancing discourse coherence.

\subsection{Baselines}
\label{app:baselines}
We compare the proposed GRAFT framework with a diverse set of baselines, ranging from commercial translation systems to advanced document-level machine translation models. Below, we describe each baseline and its corresponding setup:

\paragraph{Commercial Translation System: Google Translate API}  
We use the Google Translate API for document translation.\footnote{\url{https://py-googletrans.readthedocs.io/en/latest/}} The entire document is translated in a single request, leveraging Google's production-grade translation system.

\paragraph{Sentence-Level Baseline: NLLB}  
The NLLB-200 3.3B model \citep{nllb200} serves as our sentence-level baseline.\footnote{\url{https://huggingface.co/facebook/nllb-200-3.3B}} We use the pre-trained checkpoint to translate each document sentence by sentence, without considering inter-sentential context. This setup provides insight into the impact of ignoring document-level context on translation quality.

\paragraph{G-Trans (BART)}  
We include the G-Trans model \citep{bao-etal-2021-g}, which utilizes a BART-based architecture for document-level machine translation. The G-Trans approach employs a hierarchical attention mechanism to model inter-sentential dependencies, processing the document as a sequence of sentence representations. We adapt their training setup and evaluation protocols for our experiments.

\paragraph{GPT Models}  
We evaluate two GPT-based models, GPT-3.5-Turbo\footnote{\url{https://platform.openai.com/docs/models/gpt-3.5-turbo}} and GPT-4o-mini,\footnote{\url{https://platform.openai.com/docs/models/gpt-4o-mini}} as baselines. These models are prompted to translate the entire document from the source to the target language in one go. This setup assesses the capability of general-purpose large language models to perform document-level translation.

\paragraph{DelTA}  
We include DelTA \citep{wang2025delta}, a document-level translation agent based on multi-level memory. The model employs Qwen2-7B-Instruct and Qwen2-72B-Instruct backbones. For evaluation, we directly adopt their methodology and utilize their reported results on the TED tst2017 dataset from the IWSLT2017 translation task \citep{cettolo2012wit3}. The multi-level memory mechanism in DelTA provides a strong baseline for maintaining document coherence and consistency.

\section{Results: Design Choices for GRAFT}
\label{app:results_design_choices_for_graft}

We present the results of comparing different strategies for each of the $\mathcal{L}_{\mathrm{seg}}$, $\mathcal{L}_{\mathrm{edge}}$, and $\mathcal{L}_{\mathrm{mem}}$ as discussed in Section~\ref{subsec:analysis_of_source_document_to_dag}.

\subsection{Comparison of discourse segmentation strategies}
\label{app:discourse_segmentation_comparison_results}
Table~\ref{tab:discourse_agent_ablation} shows the performance of different discourse segmentation strategies we experimented with.

\begin{table}[h]
\centering
\begin{tabular}{crrr}
\hline
\textbf{Language Pair} & \multicolumn{1}{c}{\textbf{RS}} & \multicolumn{1}{c}{\textbf{SC}} & \multicolumn{1}{c}{\textbf{DA (Ours)}} \\ \hline
\multicolumn{4}{c}{\textbf{English $\Rightarrow$ Xx}}                                                                                         \\ \hdashline
En $\Rightarrow$ Zh    & 25.7                            & 21.8                            & \textcolor{blue}{\textbf{26.6}}                                   \\
En $\Rightarrow$ De    & 11.6                            & 24.9                            & \textcolor{blue}{\textbf{26.1}}                                   \\
En $\Rightarrow$ Fr    & 21.3                            & 33.3                            & \textcolor{blue}{\textbf{35.8}}                                   \\
En $\Rightarrow$ Ja    & 6.1                             & 6.4                             & \textcolor{blue}{\textbf{7.2}}                                    \\ \hline \hline
\multicolumn{4}{c}{\textbf{Xx $\Rightarrow$ English}}                                                                                         \\ \hdashline
Zh $\Rightarrow$ En    & 11.4                            & 23.4                            & \textcolor{blue}{\textbf{24.6}}                                   \\
De $\Rightarrow$ En    & 18.5                            & 27.9                            & \textcolor{blue}{\textbf{31.5}}                                   \\
Fr $\Rightarrow$ En    & 12.9                            & 31.3                            & \textcolor{blue}{\textbf{41.0}}                                   \\
Ja $\Rightarrow$ En    & 8.7                             & 9.5                             & \textcolor{blue}{\textbf{14.5}}                                   \\ \hline \hline
Ave.                & 14.5                            & 22.3                            & \textcolor{blue}{\textbf{25.9}}                                   \\ \hline
\end{tabular}
\caption{d-BLEU scores comparing various \textbf{discourse segmentation} strategies: Random Segmentation (RS), Semantic Similarity-based segmentation (SC), and Discourse Agent (DA).}
\label{tab:discourse_agent_ablation}
\end{table}

\subsection{Comparison of dependency modelling strategies}
\label{app:dependency_modelling_comparison_results}
Table~\ref{tab:edge_agent_ablation} shows the performance of different dependency modelling strategies we experimented with.

\begin{table}[h]
\centering
\begin{tabular}{crrr}
\hline
\textbf{Language Pair} & \multicolumn{1}{c}{\textbf{CG}} & \multicolumn{1}{c}{\textbf{TF-IDF}} & \multicolumn{1}{c}{\textbf{EA (Ours)}} \\ \hline
\multicolumn{4}{c}{\textbf{English $\Rightarrow$ Xx}}                                                                                             \\ \hdashline
En $\Rightarrow$ Zh    & 24.3                            & 25.4                                & \textcolor{blue}{\textbf{26.6}}                                   \\
En $\Rightarrow$ De    & 25.2                            & 23.0                                & \textcolor{blue}{\textbf{26.1}}                                   \\
En $\Rightarrow$ Fr    & 35.7                            & 31.2                                & \textcolor{blue}{\textbf{35.8}}                                   \\
En $\Rightarrow$ Ja    & 4.7                             & 5.9                                 & \textcolor{blue}{\textbf{7.2}}                                    \\ \hline \hline
\multicolumn{4}{c}{\textbf{Xx $\Rightarrow$ English}}                                                                                             \\ \hdashline
Zh $\Rightarrow$ En    & \textcolor{blue}{\textbf{24.8}}                            & 23.1                                & 24.6                                   \\
De $\Rightarrow$ En    & 27.9                            & 28.8                                & \textcolor{blue}{\textbf{31.5}}                                   \\
Fr $\Rightarrow$ En    & 37.0                            & 39.8                                & \textcolor{blue}{\textbf{41.0}}                                   \\
Ja $\Rightarrow$ En    & 11.2                            & 13.3                                & \textcolor{blue}{\textbf{14.5}}                                   \\ \hline \hline
Ave.                & 23.9                            & 23.8                                & \textcolor{blue}{\textbf{25.9}}                                   \\ \hline
\end{tabular}
\caption{d-BLEU scores comparing various \textbf{edge construction} strategies: Chain Graph (CG), TF-IDF based edge construction (TF-IDF), and Edge Agent (EA).}
\label{tab:edge_agent_ablation}
\end{table}

\subsection{\texorpdfstring{Effects of Local Memory ($\mathrm{mem}_i$) components}{Effects of Local Memory}}
\label{app:contribution_of_memory_components}
Table~\ref{tab:memory_agent_ablation} shows the performance of GRAFT by ablating each memory component.

\begin{table*}[h]
\centering
\begin{tabular}{crrrrrrr}
\hline
    \textbf{Language Pair} & \multicolumn{1}{l}{\textbf{NM}} & \multicolumn{1}{l}{\textbf{w/o $M_i^{\mathrm{np}}$}} & \multicolumn{1}{l}{\textbf{w/o $M_i^{\mathrm{ent}}$}} & \multicolumn{1}{l}{\textbf{w/o $C_i$}} & \multicolumn{1}{l}{\textbf{w/o $M_i^{\mathrm{phr}}$}} & \multicolumn{1}{l}{\textbf{w/o $S_i$}} & \multicolumn{1}{l}{\textbf{FM}} \\ \hline
\multicolumn{8}{c}{\textbf{English $\Rightarrow$ Xx}}                                                                                                                                                                                                                                                                                                         \\ \hdashline
En $\Rightarrow$ Zh    & 21.5                            & 26.2                                                 & 23.3                                                  & 24.6                                   & 23.6                                                  & 25.9                                   & \textcolor{blue}{\textbf{26.6}}                            \\
En $\Rightarrow$ De    & 23.5                            & 25.4                                                 & 22.8                                                  & 25.9                                   & 21.4                                                  & 24.1                                   & \textcolor{blue}{\textbf{26.1}}                            \\
En $\Rightarrow$ Fr    & 27.9                            & 35.1                                                 & 24.1                                                  & 34.6                                   & 28.6                                                  & 34.5                                   & \textcolor{blue}{\textbf{35.8}}                            \\
En $\Rightarrow$ Ja    & 3.8                             & 5.6                                                  & 5.9                                                   & 6.0                                    & 5.4                                                   & 6.5                                    & \textcolor{blue}{\textbf{7.2}}                             \\ \hline \hline
\multicolumn{8}{c}{\textbf{Xx $\Rightarrow$ English}}                                                                                                                                                                                                                                                                                                         \\ \hdashline
Zh $\Rightarrow$ En    & 16.3                            & 24.9                                                 & 23.2                                                  & 23.7                                   & 25.2                                                  & 23.8                                   & \textcolor{blue}{\textbf{24.6}}                            \\
De $\Rightarrow$ En    & 23.2                            & 26.6                                                 & 30.7                                                  & 29.0                                   & 30.5                                                  & 29.8                                   & \textcolor{blue}{\textbf{31.5}}                            \\
Fr $\Rightarrow$ En    & 35.3                            & 40.6                                                 & 37.7                                                  & 36.0                                   & 31.7                                                  & 40.0                                   & \textcolor{blue}{\textbf{41.0}}                            \\
Ja $\Rightarrow$ En    & 9.8                             & 14.0                                                 & 13.1                                                  & 9.6                                    & 12.8                                                  & 12.6                                   & \textcolor{blue}{\textbf{14.5}}                            \\ \hline \hline
Ave.                & 20.2                            & 24.8                                                 & 22.6                                                  & 23.7                                   & 22.4                                                  & 24.6                                   & \textcolor{blue}{\textbf{25.9}}                            \\ \hline
\end{tabular}
\caption{Ablation on the components of \textbf{memory}, with NM denoting no use of memory, and FM denoting use of all the memory components.}
\label{tab:memory_agent_ablation}
\end{table*}

\section{Analysis: Plots}
\label{app:plots}

\begin{figure}[h]
    \centering
    \includegraphics[width=0.85\columnwidth]{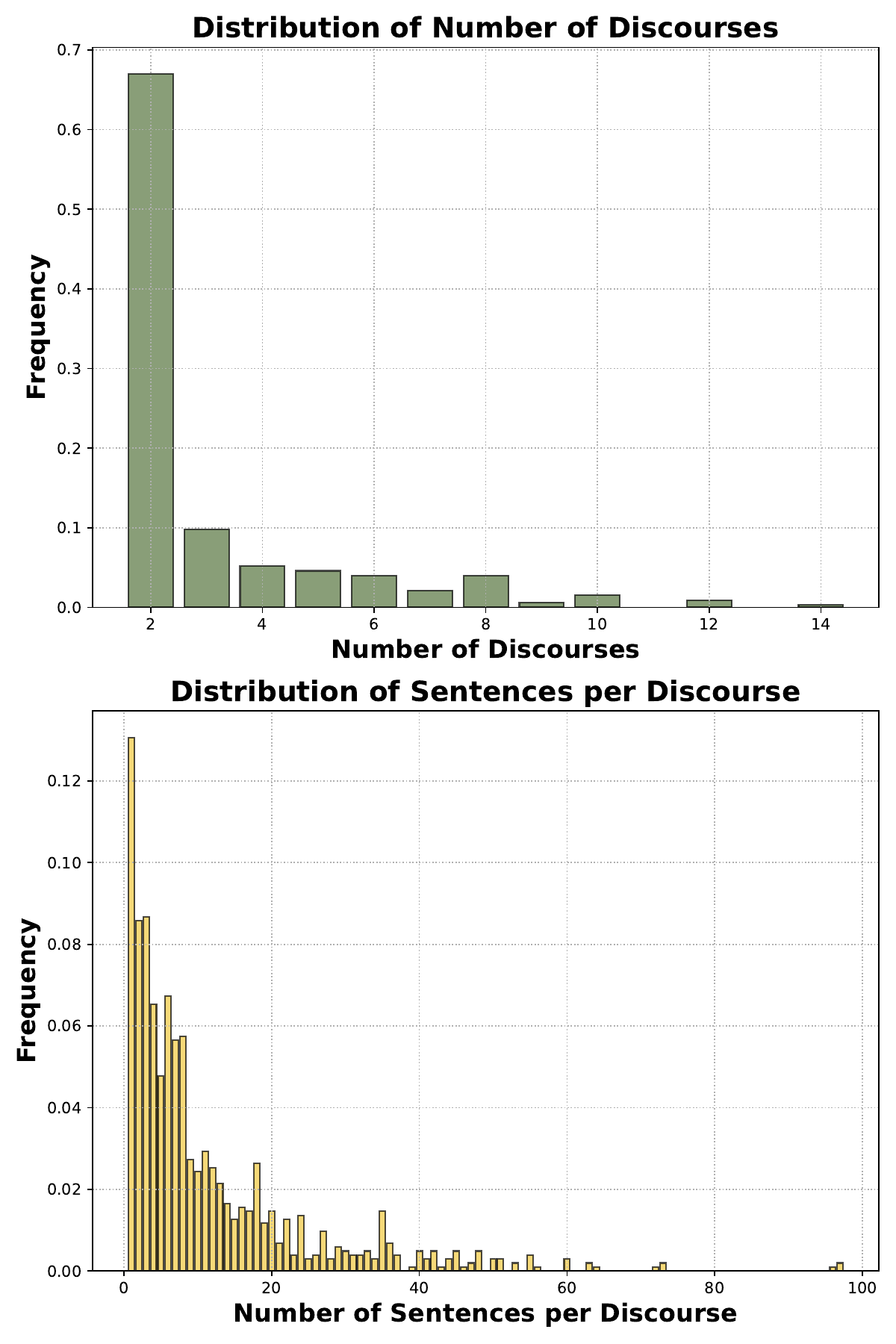}
    \caption{Analysis of the Discourse Agent with distribution of frequency of number of discourses (per document) and distribution of sentences in each discourse.}
    \label{fig:discourse_agent_analysis}
\end{figure}

\begin{figure}[h]
    \centering
    \includegraphics[width=0.85\columnwidth]{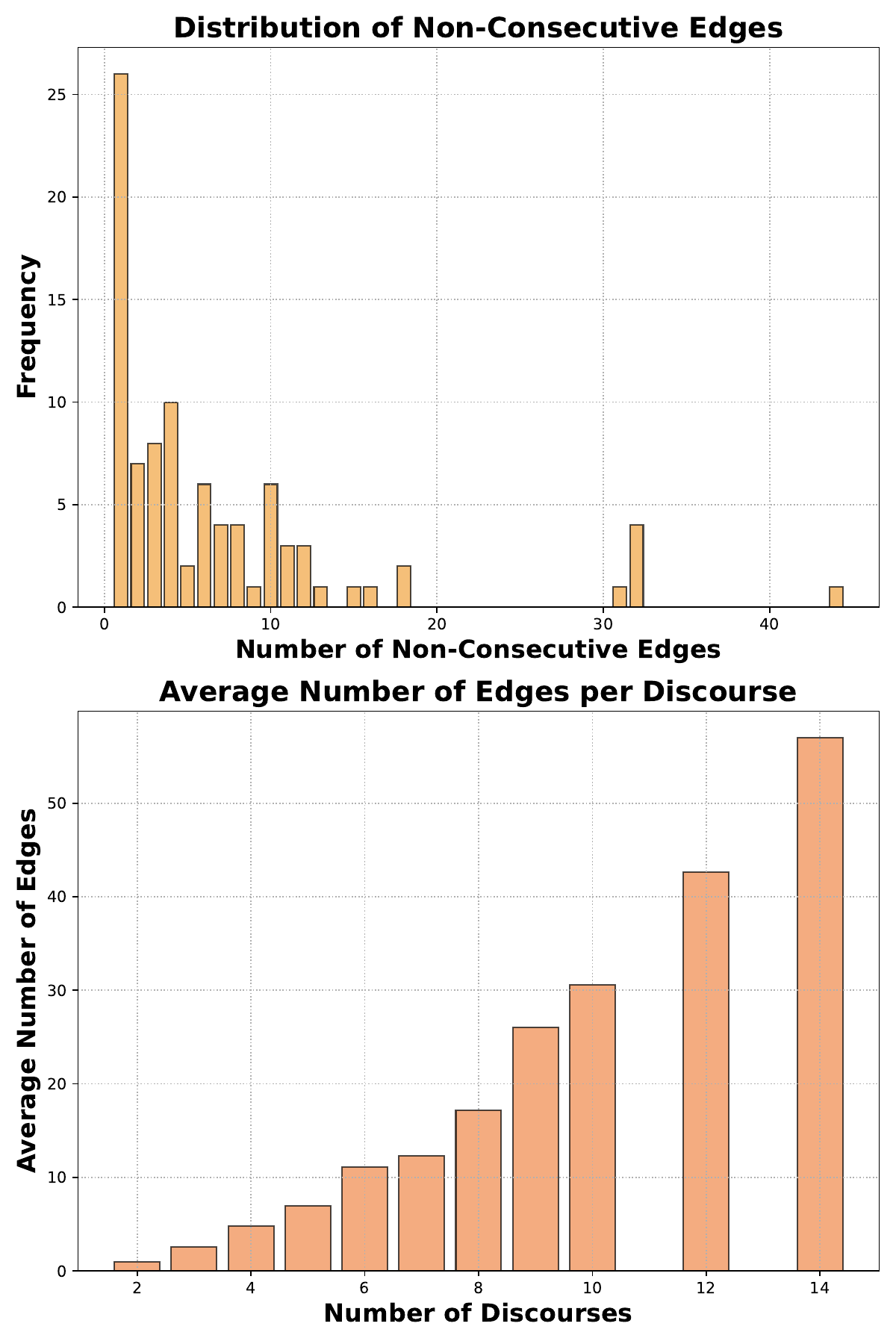}
    \caption{Evaluation of the Edge Agent with distribution of non-consecutive edges over documents, and distribution of number of edges over discourses.}
    \label{fig:edge_agent_analysis}
\end{figure}

\begin{figure}[h]
    \centering
    \includegraphics[width=0.85\columnwidth]{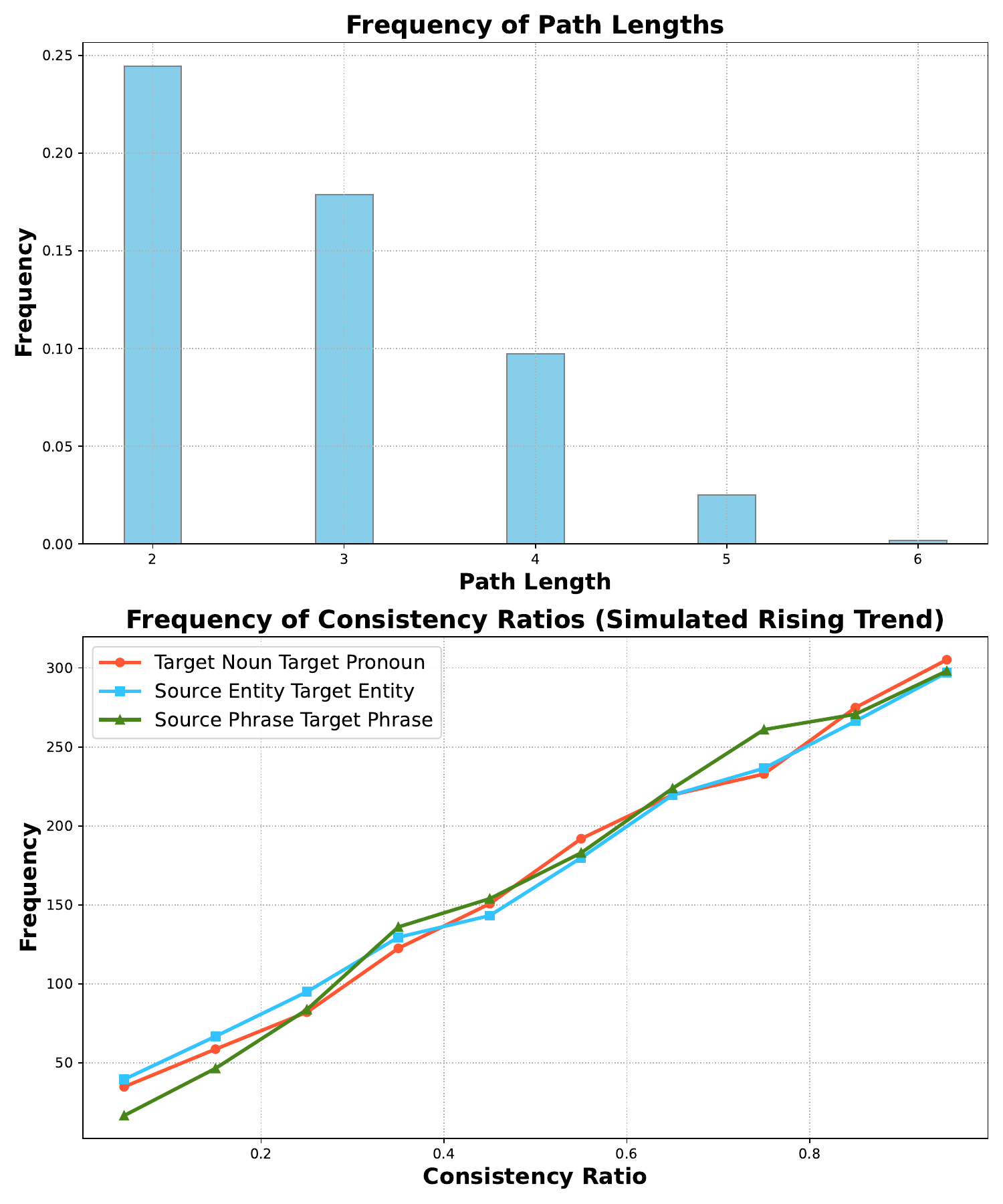}
    \caption{Assessment of system consistency through two measures: intra-document coherence and inter-document fidelity. Results demonstrate the effectiveness of system-level enhancements in ensuring stable outputs.}
    \label{fig:consistency_analysis}
\end{figure}

The plots for discussion in Section~\ref{sec:analysis}. Figure~\ref{fig:discourse_agent_analysis} shows the distribution of the number of discourse segments per document, as well as the number of sentences grouped within a discourse. Similarly, Figure~\ref{fig:edge_agent_analysis} illustrates the distribution of the count of edges per document and the trend of edge counts as a function of discourse segments. Figure~\ref{fig:consistency_analysis} shows the distribution of the path lengths and the consistency ratio.

\section{Human Evaluation: Annotator Details and Guidelines}
\label{app:human_evaluation_guidelines} 

The evaluation process was conducted by human annotators with an average age of 24 years. All annotators hold a Bachelor of Arts and Master of Arts degree in Chinese Language and Literature. Each annotator received a compensation of \$200 for their participation. The evaluation process spanned 3 to 4 days, during which each annotator performed the evaluation independently. For all human evaluations conducted, we report the average performance based on their assessments. The guidelines for human evaluation have been provided in Table~\ref{tab:quality-discourse}.

\begin{table*}[ht]
\centering
\begin{tabular}{|c|p{6.5cm}|p{6.5cm}|}
\hline
\textbf{Score} & \textbf{General Quality} & \textbf{Discourse Awareness} \\
\hline
5 & 
Translation passes quality control; the overall translation is excellent. Translation is very fluent with no grammatical errors and has been localized to fit the target language. Word choice is accurate with no mistranslations. The translation is 100\% true to the source text.
& 
No inconsistency relating to key terms such as names, organization, etc. Linking words or expressions between sentences keeps the logic and language of the passage clear and fluent. Context and tone are consistent throughout. The style conforms to the culture and habits of the target language. \\
\hline
4 & 
Translation passes quality control; the overall translation is very good. Translation is fluent. Any errors that may be present do not affect the meaning or comprehension of the text. Most word choices are accurate, but some may cause ambiguity. Key terms are consistent. Inconsistency is limited to non-key terms.
& 
Logical and language is clear and fluent. Some sentences lack transitions but do not affect contextual comprehension. The topic is consistent. Tone and word choice may be inconsistent, but comprehension is not affected. Translation conforms to the culture and habits. \\
\hline
3 & 
Translation passes quality control; the overall translation is ok. Translation is mostly fluent, but there are many sections that require rereading due to language usage. Some word choices are inaccurate or contain errors, but the meaning of the sentence can be inferred from context.
& 
Some key terms may be inconsistent. Most sentences translate smoothly and logically, but some sentences may seem abrupt due to a lack of linkage. The topic is consistent. Tone and word choice are inconsistent, noticeably affecting the accuracy of reading comprehension. \\
\hline
2 & 
Translation does not pass quality control; the overall translation is poor. Meaning is unclear or disjointed. Even with multiple rereading, the passage may still be incomprehensible. Translation is not accurate to the source text or is missing in large quantities, causing the translation to deviate from the source text.
& 
Many key terms are inconsistent, needing multiple rereading to understand the context of the passage. Some linkages are present, but overall, the passage lacks fluency and clarity, causing trouble with comprehension. The topic or tone is different from other passages, affecting reading comprehension. \\
\hline
1 & 
Translation does not pass quality control; the overall translation is very poor. More than half of the translation is mistranslated or missing.
& 
Key terms are inconsistent, causing great trouble with comprehension. Some linkages are present, but overall, the passage lacks fluency and clarity, heavily interfering with comprehension. The topic or tone is different from other passages, heavily interfering with comprehension. \\
\hline
0 & 
Translation output is unrelated to the source text.
& 
The output is unrelated to the previous or following sections. \\
\hline
\end{tabular}
\caption{Quality and Discourse Awareness Scoring Guidelines}
\label{tab:quality-discourse}
\end{table*}

\section{Human Evaluation of GRAFT}
\label{app:human_evaluation}

To evaluate the performance of our GRAFT system, we conducted a comprehensive human evaluation (as referred in Section~\ref{sec:experiments_results}) consisting of two primary analyses: (1) Discourse Agent analysis and (2) Edge Agent analysis. These evaluations involved human annotators assessing the quality of discourse segmentation and inter-discourse dependencies, respectively, with respect to translation quality.

\subsection{Discourse Agent Analysis}
The Discourse Agent segments documents into meaningful units (discourses) intended to provide sufficient context for intra-discourse phenomena during translation. Evaluators analyzed the effectiveness of this segmentation across several phenomena, including pronoun resolution, discourse coherence, and consistency in tense and aspect.

The results indicate that $\textcolor{blue2}{\mathbf{70.4\%}}$ of pronouns were resolved correctly within the segmented discourse, demonstrating that the context provided by the Discourse Agent is effective in handling referential relationships. Additionally, the translated discourses exhibit a coherence score of $\textcolor{blue2}{\mathbf{90.2\%}}$ and consistency in tense and aspect at $\textcolor{blue2}{\mathbf{88.6\%}}$. These findings highlight the robustness of the Discourse Agent in preserving local linguistic phenomena and ensuring high-quality translations at the discourse level.

\subsection{Edge Agent Analysis}
The Edge Agent identifies dependencies between non-contiguous discourses, crucial for capturing inter-discourse phenomena. Annotators evaluated whether the edges correctly indicated dependencies and whether the memory agent successfully handled terminology consistency.

The evaluation reveals that $\textcolor{blue2}{\mathbf{76.3\%}}$ of edges are accurately identified, affirming the Edge Agent's capability to model contextual dependencies effectively. Furthermore, the integration of the memory agent allowed GRAFT to maintain terminology consistency in $\textcolor{blue2}{\mathbf{84.5\%}}$ of cases, a significant improvement over baseline systems lacking such contextual memory mechanisms.

\subsection{Domain-specific Translation: Human Evaluation}
\label{app:domain_specific_translation_human_evaluation}

Here we present the human evaluation results for our Domain-specific Translation experiment (Section~\ref{subsec:domain_specific_translation}). Table~\ref{tab:domain_results_human} presents the results of human evaluation of various systems across the four domains: News, Social, Fiction and Q\&A in Zh $\Rightarrow$ En direction. The guidelines for the human evaluation are provided in Appendix~\ref{app:human_evaluation_guidelines}.

\subsection{Implications and Observations}
These results demonstrate that the GRAFT pipeline successfully balances intra- and inter-discourse phenomena, offering robust support for document-level machine translation. The high coherence and consistency scores in the Discourse Agent analysis indicate that discourse segmentation effectively captures localized linguistic phenomena. Similarly, the accurate identification of edges and terminology consistency underscores the Edge Agent's ability to model and leverage cross-discourse dependencies.

Overall, the evaluation results validate the design of GRAFT, highlighting its potential for improving translation quality by systematically addressing discourse-level challenges. Future work may focus on further enhancing dependency modelling and exploring additional applications of memory mechanisms to strengthen cross-discourse coherence.

\begin{table*}[h]
\centering
\begin{tabular}{llllllcl}
\hline
\multirow{2}{*}{\textbf{System}} & \multicolumn{4}{c}{\textbf{Human (\textcolor{purple2}{General}/\textcolor{green2}{Discourse})}}                                                             \\ \cline{2-5} \cline{8-8} 
                                    & \textbf{News} & \textbf{Social} & \textbf{Fiction} & \textbf{Q\&A} &           & \textbf{Ave.} \\ \hline
Commercial System (Google)       & 1.9/2.0       & 1.2/1.3         & 2.1/2.4          & 1.5/1.5       &           & 1.7/1.7       \\ \hdashline
GPT-3.5                          & 2.8/2.8       & 2.5/2.7         & 2.8/2.9          & 2.9/2.9       &           & 2.8/2.8       \\
GPT-4o-mini                            & 3.3/3.4       & 2.9/2.9         & 2.6/2.8          & 3.1/3.2       &           & 3.0/3.1       \\ \hdashline
GRAFT                            & \textcolor{purple2}{\textbf{4.3}}/\textcolor{green2}{\textbf{4.5}}         & \textcolor{purple2}{\textbf{3.9}}/\textcolor{green2}{\textbf{3.9}}           & \textcolor{purple2}{\textbf{3.9}}/\textcolor{green2}{\textbf{3.9}}            & \textcolor{purple2}{\textbf{4.3}}/\textcolor{green2}{\textbf{4.6}}         & \textbf{} & \textcolor{purple2}{\textbf{4.1}}/\textcolor{green2}{\textbf{4.2}}       \\ \hline
\end{tabular}
\caption{Domain-specific performance (Human Evaluation) for Zh $\Rightarrow$ En translation direction across four domains.}
\label{tab:domain_results_human}
\end{table*}

\section{Latency and Cost Analysis of GRAFT}
\label{app:latency_and_cost_analysis}

In this section, we analyze the latency and monetary cost of GRAFT (as referred to in Section~\ref{sec:experiments_results}), comparing it to large-scale models like GPT-4o-mini and GPT-3.5-Turbo, which exhibit similar performance levels. While GRAFT delivers superior handling of intra- and inter-discourse phenomena, its agentic design leads to higher latency and cost due to the need for multiple LLM calls during the translation process.

GRAFT demonstrates substantial performance gains, with improvements of $\textcolor{blue}{\mathbf{2.0}}$ d-BLEU over GPT-4o-mini and $\textcolor{blue}{\mathbf{3.2}}$ d-BLEU over GPT-3.5-Turbo. However, this performance comes at the cost of increased translation time. On average, GRAFT takes approximately three times longer to translate a document compared to GPT-4o-mini, as its workflow involves sequential calls for discourse segmentation, edge construction, and contextual refinement. The reliance on inter-agent communication and repeated querying of LLMs makes the system inherently slower than monolithic models.

The monetary cost of GRAFT also reflects its resource-intensive nature. For a typical 1,000-word document, the estimated cost of using GRAFT is $\$0.12$, compared to $\$0.08$ for GPT-4o-mini and $\$0.05$ for GPT-3.5-Turbo. These estimates are based on API pricing for LLM usage and account for the increased number of calls made by GRAFT. The additional cost is attributed to the multi-agent structure, which requires separate LLM invocations for tasks such as discourse segmentation, dependency modelling, and translation generation.

Despite these trade-offs, GRAFT's improvements in translation quality make it a compelling choice for applications where high precision and contextual accuracy are critical. For cost- or latency-sensitive scenarios, optimizing GRAFT's architecture could help strike a balance. Potential solutions include reducing the number of LLM calls through caching mechanisms, using smaller LLMs for certain agents, or fine-tuning specific components to reduce reliance on external APIs.

Overall, this analysis highlights the trade-offs between quality, latency, and cost in GRAFT. While it excels in translation accuracy and contextual coherence, its higher operational overhead underscores the need for targeted optimizations to improve its practicality in broader use cases.

\section{Task Description and Few-shot Prompt Template}
\label{app:task_description_and_fewshot_prompt_template}

We show the prompts for En $\Rightarrow$ De language direction.

\subsection{\texorpdfstring{$f_\mathrm{LLM}$}{Discourse Agent: Decision Function}}
\label{app:discourse_agent_prompt}

\begin{figure*}[h]
    \centering
    \includegraphics[width=1\textwidth]{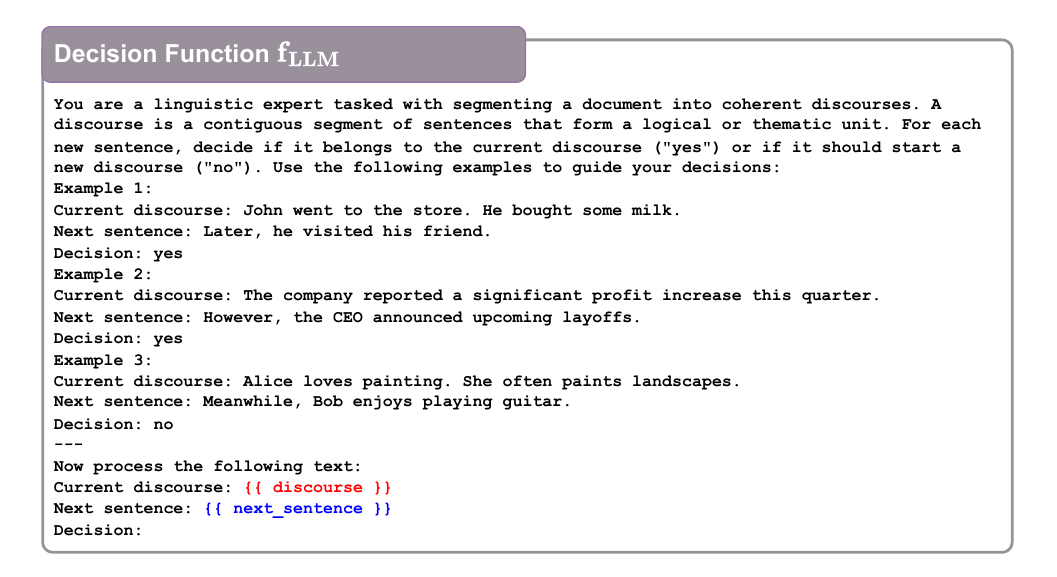}
    \caption{Task description and prompt: Decision function ($f_{\mathrm{LLM}}$) of the Discourse Agent.}
    \label{fig:prompt_discourse_agent}
\end{figure*}

The task description and prompt used for $f_\mathrm{LLM}$ are shown in Figure~\ref{fig:prompt_discourse_agent}.

\subsection{\texorpdfstring{$\mathcal{L}_{\mathrm{edge}}$}{Edge Agent}}
\label{app:edge_agent_prompt}
\begin{figure*}[h]
    \centering
    \includegraphics[width=1\textwidth]{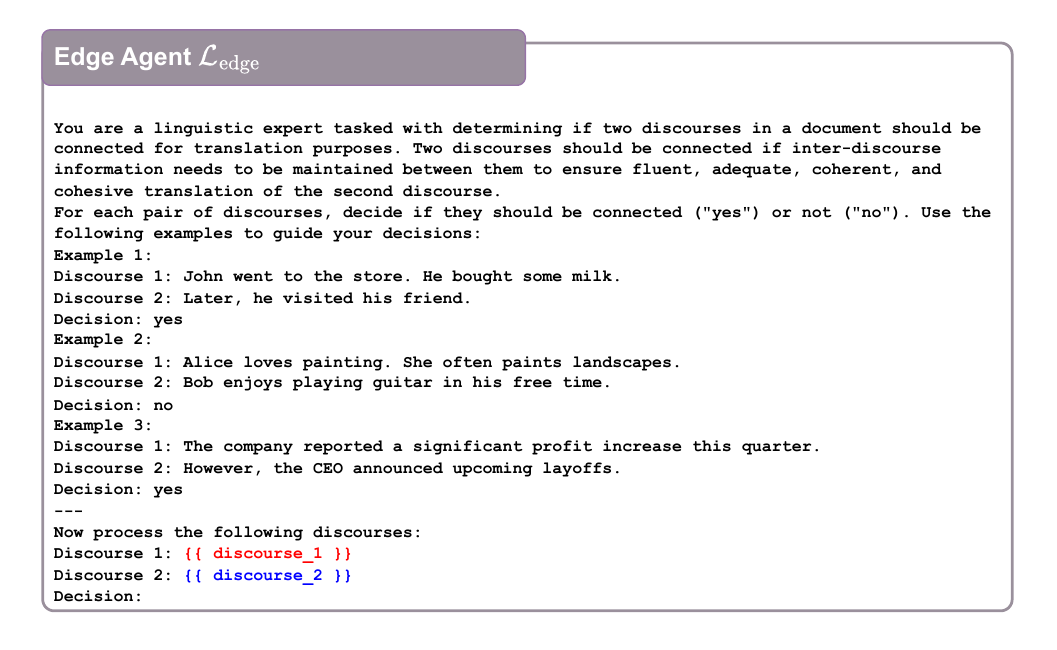}
    \caption{Task description and prompt: Edge Agent ($\mathcal{L}_{\mathrm{edge}}$).}
    \label{fig:prompt_edge_agent}
\end{figure*}

The task description and prompt used for $\mathcal{L}_{\mathrm{edge}}$ is shown in Figure~\ref{fig:prompt_edge_agent}.

\subsection{\texorpdfstring{$\mathcal{L}_{\mathrm{mem}}$}{Memory Agent}}
\label{app:memory_agent_prompt}
\begin{figure*}[h]
    \centering
    \includegraphics[width=1\textwidth]{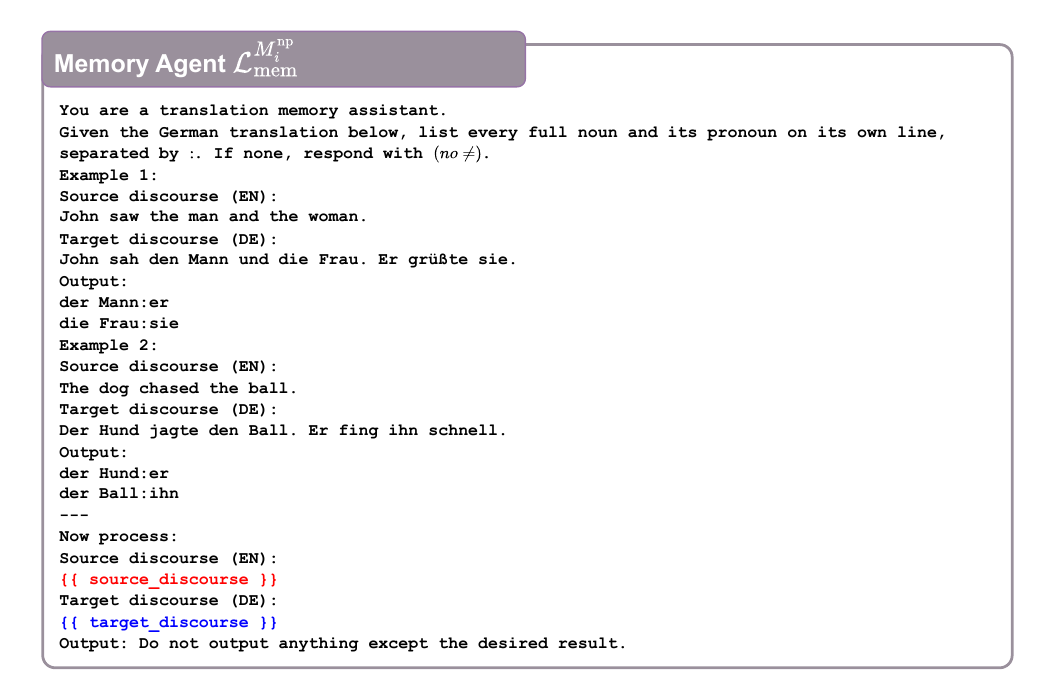}
    \caption{Task description and prompt: Memory Agent: $M_i^{\mathrm{np}}$}
    \label{fig:prompt_memory_agent_target_noun_target_pronoun}
\end{figure*}

\begin{figure*}[h]
    \centering
    \includegraphics[width=1\textwidth]{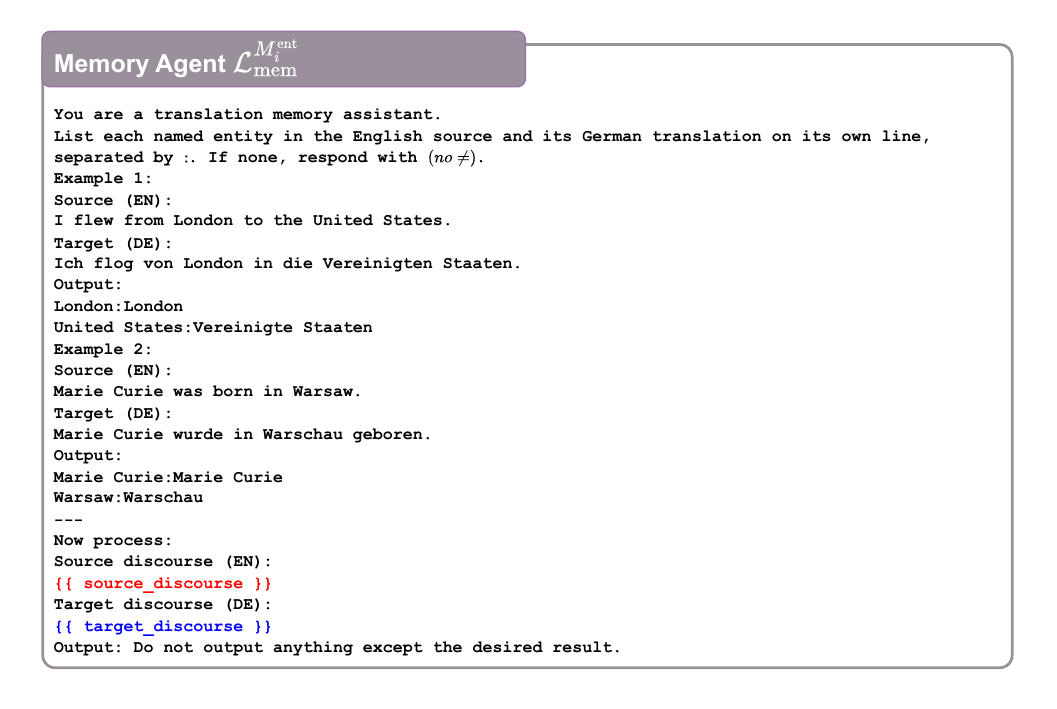}
    \caption{Task description and prompt: Memory Agent: $M_i^{\mathrm{ent}}$}
    \label{fig:prompt_memory_agent_source_entity_target_entity}
\end{figure*}

\begin{figure*}[h]
    \centering
    \includegraphics[width=1\textwidth]{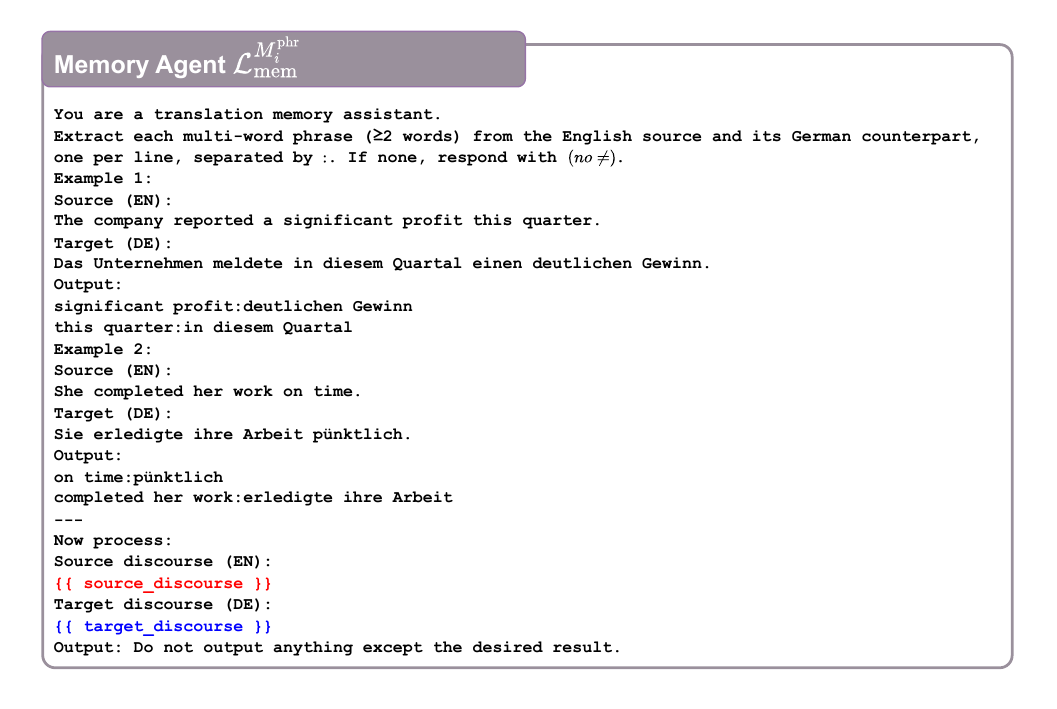}
    \caption{Prompt: Memory Agent: $M_i^{\mathrm{phr}}$}
    \label{fig:prompt_memory_agent_source_phrase_target_phrase}
\end{figure*}

\begin{figure*}[h]
    \centering
    \includegraphics[width=1\textwidth]{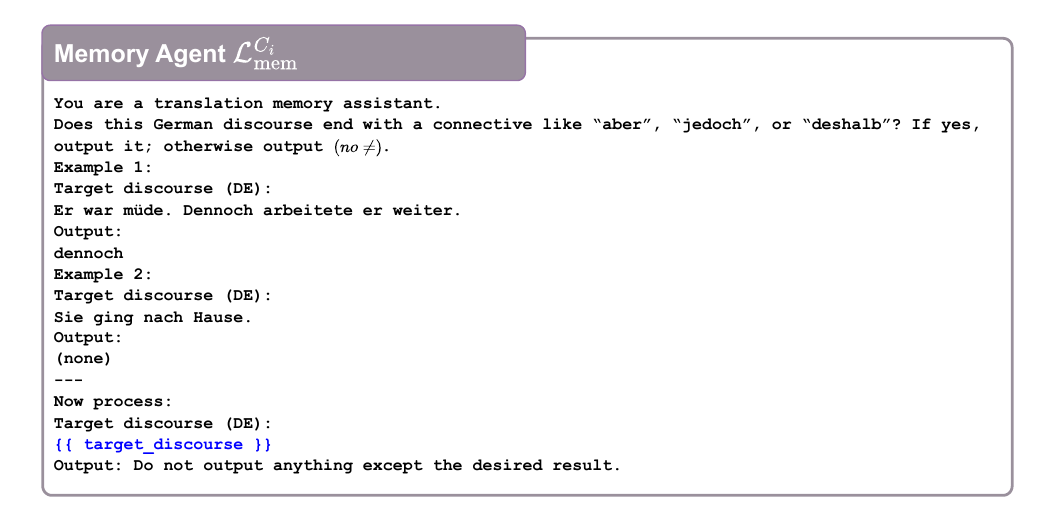}
    \caption{Task description and prompt: Memory Agent: $C_i$}
    \label{fig:prompt_memory_agent_discourse_connectives}
\end{figure*}

\begin{figure*}[h]
    \centering
    \includegraphics[width=1\textwidth]{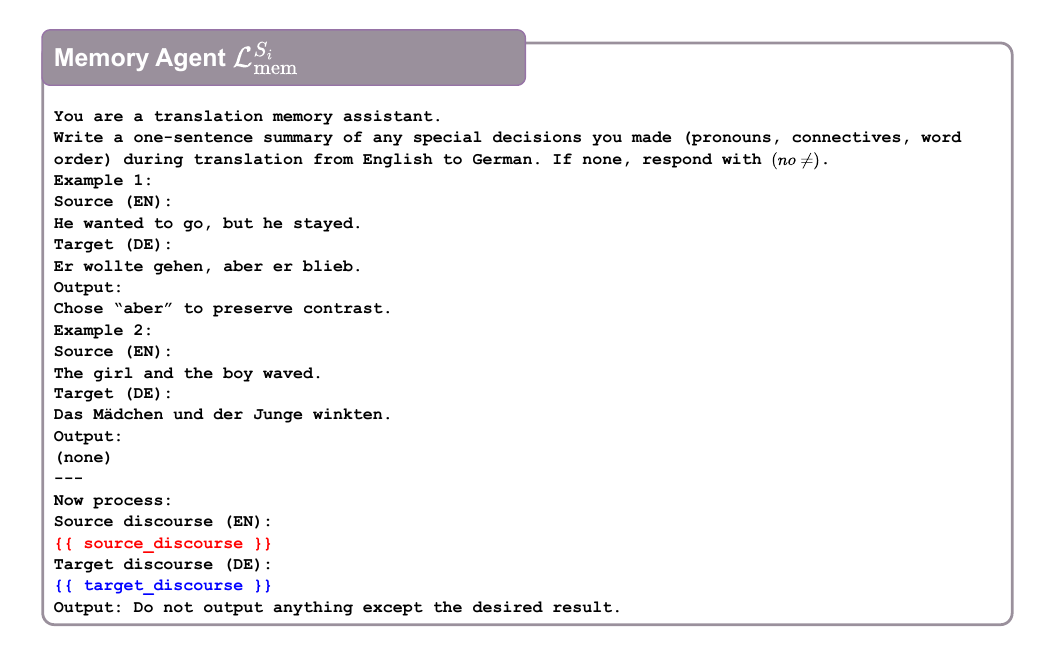}
    \caption{Task description and prompt: Memory Agent: $S_i$}
    \label{fig:prompt_memory_agent_summary}
\end{figure*}

The task description and prompt used to extract various components of the memory are shown in Figure~\ref{fig:prompt_memory_agent_target_noun_target_pronoun}, Figure~\ref{fig:prompt_memory_agent_source_entity_target_entity}, Figure~\ref{fig:prompt_memory_agent_source_phrase_target_phrase}, Figure~\ref{fig:prompt_memory_agent_discourse_connectives}, and Figure~\ref{fig:prompt_memory_agent_summary}.

\subsection{\texorpdfstring{$\mathcal{L}_{\mathrm{trans}}$}{Translation Agent}}
\label{app:translation_agent_prompt}
\begin{figure*}[h]
    \centering
    \includegraphics[width=1\textwidth]{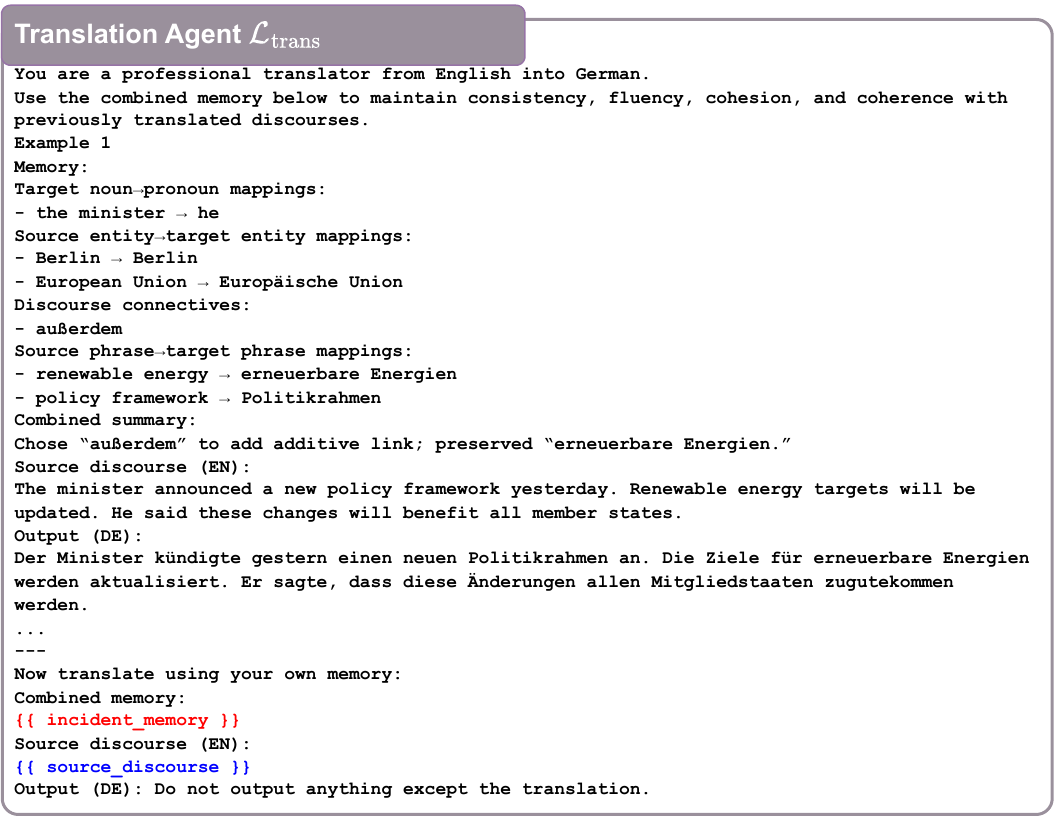}
    \caption{Task description and prompt: Translation Agent ($\mathcal{L}_{\mathrm{trans}}$)}
    \label{fig:prompt_translation_agent}
\end{figure*}

The task description and prompt used for $\mathcal{L}_{\mathrm{trans}}$ is shown in Figure~\ref{fig:prompt_translation_agent}.

\end{document}